\documentclass[10pt,final,journal,twocolumn]{IEEEtran}
\usepackage[caption=false,font=footnotesize]{subfig}
\usepackage{algorithm}
\usepackage{algorithmic}
\usepackage{graphicx}
\usepackage{amsmath,amsthm,amssymb}
\usepackage{bbm,dsfont}

\usepackage{hyperref}
\usepackage{multirow}
\usepackage{booktabs}
\usepackage{url}
\usepackage{color}

\graphicspath{{./graph/pdf/}}

\newtheorem{theorem}{Theorem}

\newtheorem{lemma}{Lemma}

\newtheorem{corollary}{Corollary}

\newcommand{\etal}{et al.\thinspace}

\newcommand{\eq}[1]{Eq.\thinspace\eqref{#1}}
\newcommand{\fig}[1]{Fig.\thinspace\ref{#1}}

\newcommand{\neuron}{\operatorname{neuron}}

\newcommand{\sos}{\textsc{sum-of-sum}}
\newcommand{\som}{\textsc{sum-of-max}}
\newcommand{\boe}{\textsc{bail-out-early}}

\title{A Massively Parallel Associative Memory Based on Sparse Neural Networks}
\author{
    \IEEEauthorblockN{Zhe~Yao\IEEEauthorrefmark{1}, Vincent~Gripon\IEEEauthorrefmark{2} and Michael~G.~Rabbat\IEEEauthorrefmark{1}}
    \thanks{\IEEEauthorrefmark{1}~Z.~Yao and M.G.~Rabbat are with the Department of Electrical and Computer Engineering, McGill University, Montr\'{e}al, QC, Canada. Email: \href{mailto:zhe.yao@mail.mcgill.ca}{zhe.yao@mail.mcgill.ca}, \href{mailto:michael.rabbat@mcgill.ca}{michael.rabbat@mcgill.ca}.
    }
    \thanks{\IEEEauthorrefmark{2}~V.~Gripon is with the Electronics Department, T\'{e}l\'{e}com Bretagne, Brest, France. Email: \href{mailto:vincent.gripon@telecom-bretagne.eu}{vincent.gripon@telecom-bretagne.eu}.
    }
}

\begin{document}
\maketitle
\begin{abstract}
Associative memories store content in such a way that the content can be later retrieved by presenting the memory with a small portion of the content, rather than presenting the memory with an address as in more traditional memories. Associative memories are used as building blocks for algorithms within database engines, anomaly detection systems, compression algorithms, and face recognition systems. A classical example of an associative memory is the Hopfield neural network. Recently, Gripon and Berrou have introduced an alternative construction which builds on ideas from the theory of error correcting codes and which greatly outperforms the Hopfield network in diversity, capacity and efficiency. In this paper we implement a variation of the Gripon-Berrou associative memory on a general purpose graphical processing unit (GPU). The work of Gripon and Berrou proposes two retrieval rules, \sos{} and \som{}. The \sos{} rule uses only matrix-vector multiplication and is easily implemented on the GPU. The \som{} rule is much less straightforward to implement because it involves non-linear operations. However, the \som{} rule gives significantly better retrieval error rates. We propose a hybrid rule tailored for implementation on a GPU which achieves a $880$-fold speedup without sacrificing any accuracy.
\end{abstract}


\section{Introduction}
We are all familiar with conventional memory systems where the address space and the information content stored in the memory are kept separate.
For instance, given a mailbox number, we can fetch the parcels inside, and in a modern computer, the CPU retrieves a stored integer from RAM by accessing a specified $32$- or $64$-bit hardware address.

An associative memory is a device or data structure that maps input patterns to output patterns.
It differs from conventional memory systems in that no explicit addresses are constructed.
Associative memories store paired patterns.
Then, given an input pattern, the associative memory produces the paired output pattern.
Since no explicit address is involved in its operation, the content of the input pattern itself associates directly with the paired output pattern, from which the name associative memory originates.
Although associative memories could be implemented using conventional memory systems, neural networks have been used as associative memories which retrieve patterns without having to search through the stored pattern space.
It is worth noting that hash tables, implemented using conventional memory systems, resemble associative memories since they map keys (inputs) to values (outputs), but still an explicit address needs to be generated first.

Associative memories can be categorized into two types~\cite{rajasekaran2004neural}: hetero-associative (e.g., linear associator~\cite{anderson1988neurocomputing,anderson1993neurocomputing2}, bidirectional associative memories~\cite{kosko1988bidirectional} and Willshaw networks~\cite{willshaw1969non}) and auto-associative (e.g., Hopfield networks~\cite{hopfield1982neural,hopfield1984neurons}).
Hetero-associative memories associate input with output patterns of possibly distinct nature and formats, whereas auto-associative memories are a special case where input and output patterns coincide.
This paper focuses on auto-associative memories.

Associative memories have applications in a variety of domains.
For instance, in communication networks~\cite{kaxiras2005ipstash}, routers need to quickly determine which port an incoming frame should be forwarded to based on the destination IP address.
In signal and image processing~\cite{valle2009class}, one commonly needs to match noisy or corrupted data to a predefined template.
Similar tasks appear in database engines~\cite{lin1976rares}, anomaly detection systems~\cite{bu2004camnids}, data compression algorithms~\cite{lin2000camlz}, face recognition systems~\cite{zhang2005gabor} and many other machine learning frameworks.

\subsection{Historical Background}
Associative memories have a long history within the field of neural networks.
Associative memories provide two operations: storing and retrieving.
In the storing operation, pairs of patterns are fed into the memory and the internal connections between neurons are modified, forming an aggregated representation of the stored pairs.
In the retrieving operation (also referred to as ``decoding''), the associative memory is presented with a probe pattern, which may be a corrupted or modified version of the stored pattern, and the memory should retrieve the most relevant pattern that was previously stored in a quick and reliable manner.

\begin{figure}
\centering
\includegraphics[scale=.5]{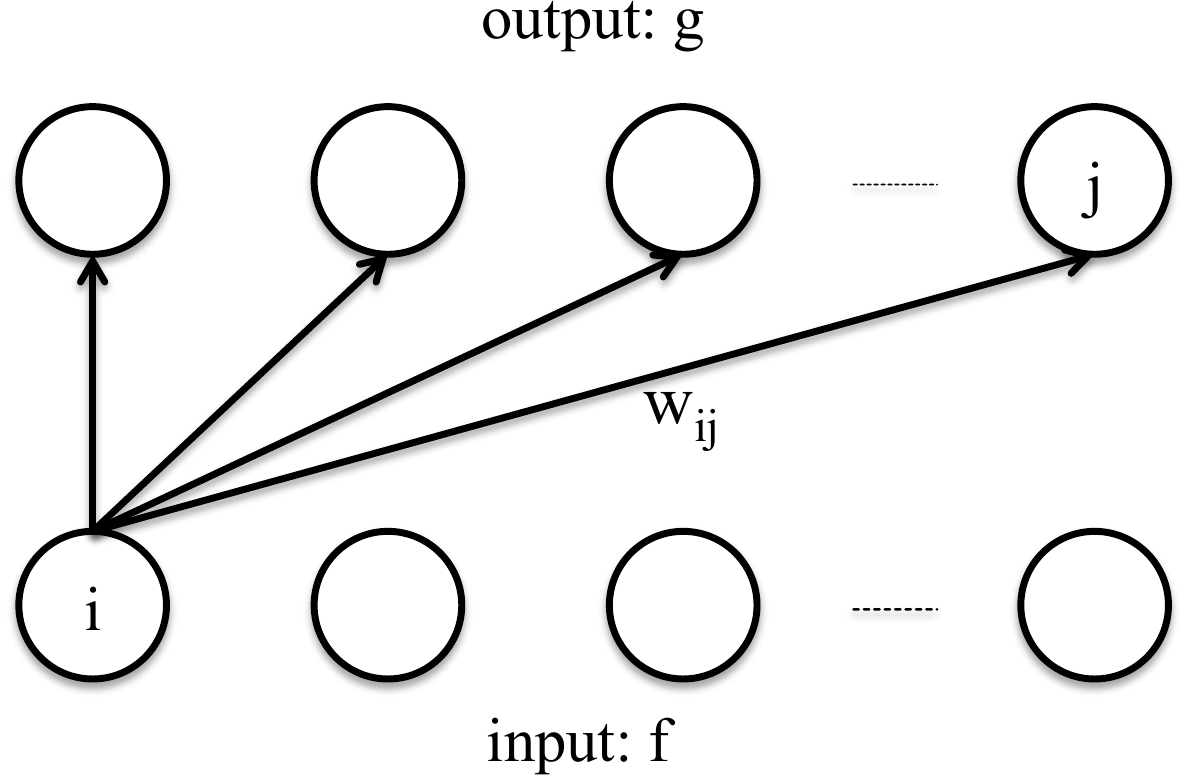}
\caption{An example of a linear associator network. Only the synapses of the first neuron in the input layer are drawn. A weight $w_{ij}$ is assigned to the synapse between neuron $i$ and neuron $j$.}
\label{fig:linearassociator}
\end{figure}

The linear associator~\cite{anderson1988neurocomputing,anderson1993neurocomputing2} is one of the simplest and earliest associative memory models; see \fig{fig:linearassociator} for an illustration.
A linear associator has an input layer and an output layer.
Synapses only exist between these two layers, hence the network can be viewed as a bipartite graph.
Connections in the network are directed from input to output neurons.
The number of neurons in each layer can be different in general, so the linear associator can be used as both an auto-associative and a hetero-associative memory.
While storing patterns, the linear associator modifies link weights according to Hebb's rule~\cite{jain1996artificial}.
While decoding a pattern, the network is presented with a given input pattern, and the paired output pattern is retrieved from the output layer immediately after one step of feed forward computation.
Since the paired pattern depends on a linear combination of the input pattern values, if all the input patterns are pairwise orthogonal, then the linear associator can reconstruct the paired patterns perfectly.
However, in most cases the orthogonality does not hold, thus the network diversity (i.e., the number of patterns that the network can store) is extremely low.

The first formal analysis of associative memories by Willshaw~\cite{willshaw1969non,willshaw1971models} dates back to early 1970's.
The structure of a Willshaw network is similar to a linear associator; it is a two-layer fully connected network, with the exception that the weights on the synapses are constrained to be 0 or 1.
The plausibility of biological neural networks discourages a fully connected network.
Therefore, Buckingham and Willshaw~\cite{buckingham1993setting} study an incomplete connected network and propose several retrieval strategies to recall the patterns.
Although simple, the Willshaw network is one of the most efficient model in terms of information stored per bit of memory (0.68 for hetero-associative and half of that for auto-associative~\cite{palm1980associative,bosch1998information}, compared to 0.14 for a Hopfield network~\cite{amit1987statistical}).
For the history and interesting developments of the Willshaw network, see the recent survey~\cite{palm2013neural} and the references therein.

\begin{figure}
\centering
\includegraphics[scale=.5]{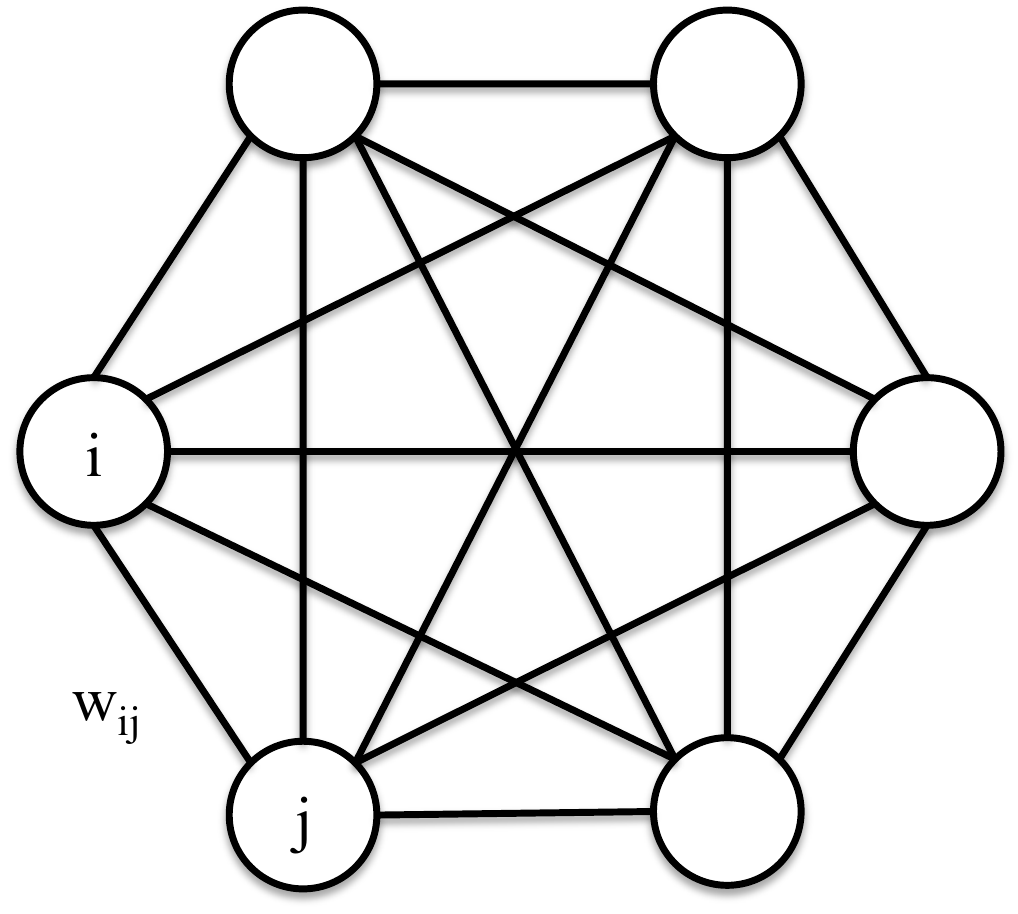}
\caption{An example of a Hopfield network with $6$ neurons. $w_{ij}$ is the weight associated with the synapse between neuron $i$ and neuron $j$.}
\label{fig:hopfield}
\end{figure}

Hopfield's seminal work~\cite{hopfield1982neural,hopfield1984neurons} on associative memories brought these structures to the attention of the neural network community in the early 1980's.
\fig{fig:hopfield} shows an example of a Hopfield network, which is a bidirectional complete graph.
Instead of having two layers, Hopfield networks comprise one layer of a bidirectional complete graph, acting as both input and output.
Therefore, it can only be used as an auto-associative memory.
Retrieval of a pattern from the network proceeds recurrently; i.e., when an impulse enters the network, the (output) values at iteration $t$ serve as the input values at iteration $t+1$, and the values iterate until the network reaches its stable configuration if it ever converges.
Kosko~\cite{kosko1988bidirectional} extends the Hopfield network into a two-layer bidirectional associative memory (BAM).
BAMs are different from linear associators because the edges in a BAM are not directed, and the retrieval rule is different.
In a BAM, values at the input and output iterate until an equilibrium is reached.
Since a BAM incorporates distinct input and output layers, it can be used as both a hetero-associative and an auto-associative memory, filling the gap between the linear associator and Hopfield networks.

\subsection{Related Work}
The recent work of Gripon and Berrou~\cite{gripon2011simple,gripon2011sparse} proposes a new family of sparse neural network architectures for associative memories.
We refer to these as \emph{Gripon-Berrou neural networks} (GBNNs).
In short, GBNNs are a variant of the Willshaw networks with a $C$-partite structure for some $C>0$.
The GBNN combines the notion of recurrence from Hopfield networks with ideas from the field of error correcting codes, and achieves nearly optimal retrieval performance.
A detailed description of the GBNN architecture and operation is given in Section~\ref{sec:gbnn}.

The GBNN is not the first attempt to link the associative memory with error correcting codes.
For example, Berrou and Gripon~\cite{berrou2010coded} successfully introduce a set of Walsh-Hadamard codes in the framework of BAMs.
The same authors also consider the use of sparse coding in a Hopfield network.
They show that, given the same amount of storage, the GBNN outperforms conventional Hopfield networks in diversity, capacity (i.e., the maximum amount of stored information in bits), and efficiency (i.e., the ratio between capacity and the amount of information in bits consumed by the network when diversity reaches its maximum), while decreasing the retrieval error.
In~\cite{gripon2012nearly}, GBNNs are interpreted using the formalism of error correcting codes, and a new retrieval rule is introduced to further decrease the error rate.
Jiang~\etal~\cite{jiang2012learning} modify the GBNN structure to learn long sequences by incorporating directed edges into the network.
Aliabadi~\etal~\cite{aliabadi2012sparse} make the extension to learn sparse messages.

The literature mentioned in the paragraphs above focuses on studying theoretical properties of GBNNs.
To be useful in many applications, it is also essential to develop fast and efficient implementations of GBNNs.
Jarollahi~\etal~\cite{jarollahi2012architecture} demonstrate a proof-of-concept implementation using the \emph{field programmable gate array} (FPGA).
Due to hardware limitations, their implementation is constrained to have at most $400$ neurons.
Larras~\etal~\cite{larras2013analog} implement an analog version of the same network which consumes $1165\times$ less energy but is $2\times$ more efficient both in the surface of the circuit and speed, compared with an equivalent digital circuit.
However, the network size is further constrained to $208$ neurons in total.

\subsection{Contributions}
The primary contribution of this paper is to demonstrate an implementation of GBNNs on a GPU using the \emph{compute unified device architecture} (CUDA).
Our massively parallel implementation supports a much larger number of neurons than existing ones, and is $880\times$ faster than a CPU implementation using optimized C++ libraries for linear algebra operations, without any loss of retrieval accuracy.
All the existing algorithms can hopefully benefit from the exciting result we present.

Towards developing an efficient parallel GBNN implementation, we study two retrieval rules: \sos{} and \som{}, which have been previously proposed in~\cite{gripon2011sparse} and \cite{gripon2012nearly}.
\sos{} is fast to implement in CUDA, because it requires only a matrix-vector multiplications, a highly optimized operation.
\som{} is slower because it involves non-linear operations, but it gives superior retrieval performance (lower error rates).
We illustrate that, although faster, \sos{} can lead to problematic oscillations.
We also prove that the \som{} rule is guaranteed to converge, and we derive properties of both rules.

The tremendous speedup mentioned above comes from two main sources.
First, we exploit the highly parallel architecture of the GPU to carry out operations efficiently.
Second, we develop a hybrid retrieval scheme using aspects of both \sos{} and \som{}, which is tailored to parallel decoding architectures.
Although we discuss a GPU implementation, we believe the ideas presented here can be used to accelerate associative memory implementations on other parallel architectures.

We emphasize that this work neither focuses on the GBNN model itself (see~\cite{gripon2011simple,gripon2011sparse}), nor carries out comparative studies with other associative memory implementations, e.g.,~\cite{austin1998advanced}.
Instead, we are interested in developing robust and fast procedures to recall stored information given incomplete probes.
As a motivating example, consider recovering messages over an erasure channel, which is a common scenario, especially in the ubiquitous IP-based communications.
The errors encountered by the IP packets can be mitigated by checksums and error correcting codes.
However, missing packets have to be recovered to avoid time-consuming and unreliable retransmissions.

\subsection{Paper Organization}
The rest of this paper is structured as follows.
Section~\ref{sec:gbnn} reviews the GBNN associative memory architecture.
Section~\ref{sec:rules} reviews the \sos{} and \som{} retrieval rules.
Section~\ref{sec:acceleration} presents the proposed acceleration techniques and discusses the customized CUDA kernel functions which implement these techniques.
Section~\ref{sec:property} provides theoretical analysis and discussion of some properties of the retrieval rules considered in this work.
Section~\ref{sec:joint} proposes the novel hybrid retrieval rule.
Section~\ref{sec:experiment} presents experimental results demonstrating the significant performance improvements obtained using GPUs.
The paper concludes in Section~\ref{sec:summary}.

\section{Gripon-Berrou Neural Networks (GBNNs)\label{sec:gbnn}}
\subsection{Structure}
A message (pattern) can be divided into a tuple of smaller symbols.
Specifically, we divide the message $M$ into $C$ symbols, $M = (m_1, m_2, \dots, m_C)$, where each symbol $m_c$ takes values in a finite set of size $L$.
For example, English words of length $10$ characters could be represented as $10$ symbols from an alphabet of size $26$; alternatively, they could be represented as $5$ symbols from an alphabet of size $26^2$.
Similarly, in an image, a symbol could correspond to the intensity of a specific pixel, or to the collective intensities of a patch of pixels.
Here we work in the abstract setting of messages and symbols defined above; precisely how the associative memory is used is application-dependent.

A GBNN~\cite{gripon2011sparse} architecture to learn such messages comprises $n = CL$ binary-valued ($0$ or $1$) neurons.
The neurons are grouped into $C$ clusters of $L$ neurons each, and edges only exist between different clusters.
A message $M = (m_1, \dots, m_C)$ is represented in the network by activating (i.e., setting to $1$) one neuron in each cluster corresponding to the value of $m_c$, and setting all other neurons to $0$.
In this way, the message is naturally encoded as a binary string of length $n$ with exactly $C$ ones.

\begin{figure}
\centering
\includegraphics[scale=.5]{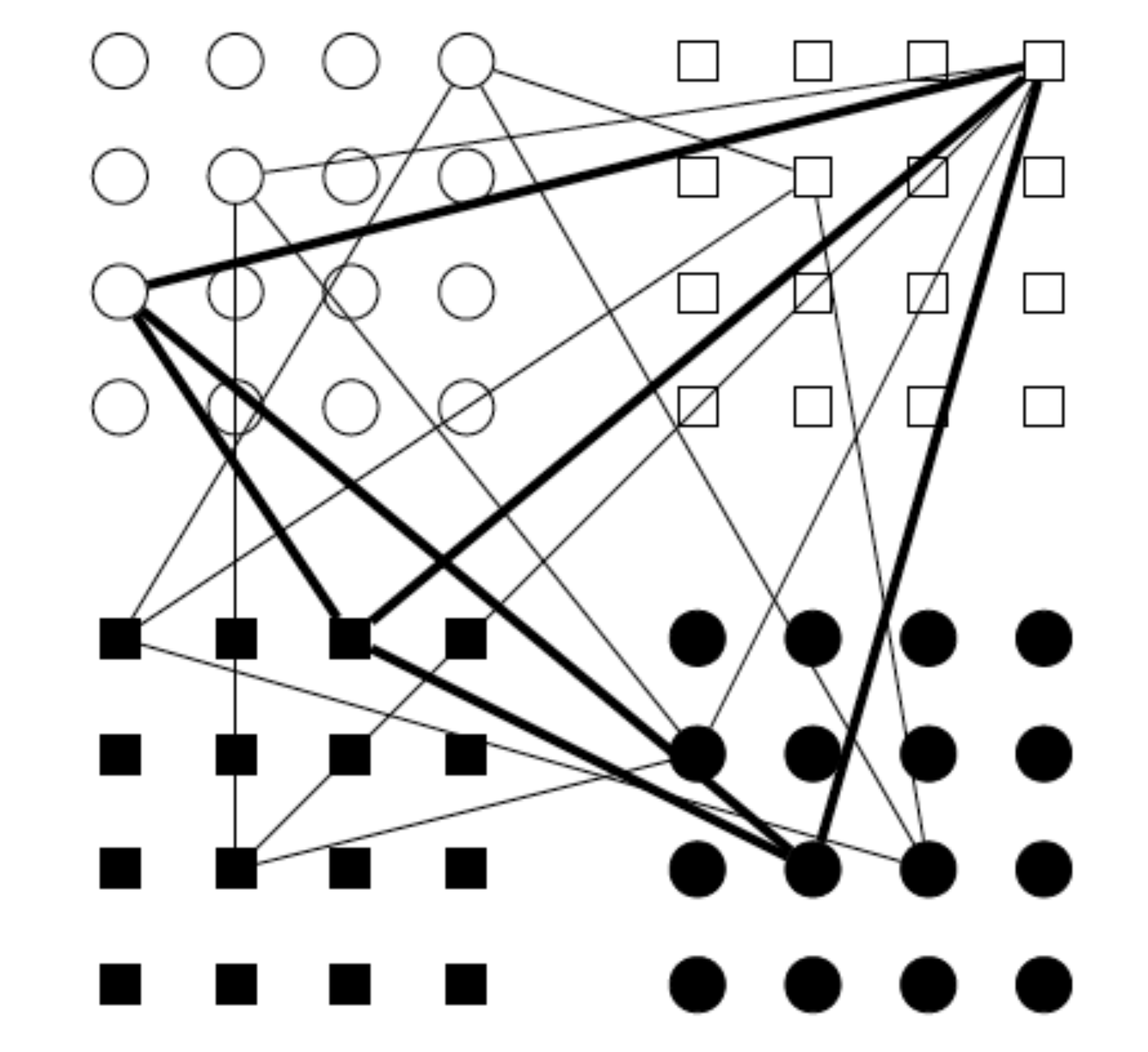}
\caption{An example of a network with $4$ clusters of $16$ neurons each~\cite{gripon2011sparse}.
We number the clusters from left to right and from top to bottom as $1 \cdots 4$.
The same scheme applies for neurons $1 \cdots 16$ within each cluster.}
\label{fig:network}
\end{figure}

When a network is initialized, all edge weights are set to zero (equivalently, there are no edges in the network).
When storing a message, we add edges to the network connecting all pairs of nodes which are activated for the particular message.
For example, consider the network depicted in \fig{fig:network}, where each message contains $C=4$ symbols and each symbol takes one of $L=16$ different values.
Let us use the convention that clusters are numbered from left to right and from top to bottom, so that $\bigcirc$'s are cluster $1$, $\Box$'s are cluster $2$, and so on; let us use the same convention within each cluster so that the neurons within each cluster are numbered from $1, 2, 3, 4$ in the first row, and so on.
The message indicated by the bold edges is $(9, 4, 3, 10)$. The edges corresponding to any single message stored in the network thus correspond to a \emph{clique}, since the neurons connected for that message form a complete sub-graph.
The binary code that represents the bold clique in \fig{fig:network} reads 0000000010000000 0001000000000000 0010000000000000 0000000001000000.

For retrieval, the network is presented with an incomplete message as a probe, e.g., $(m_1, m_2, ?, ?)$, and it must determine which (if any) stored message matches this input best.
In this paper we focus on the case where only partial messages are presented for retrieval.
If the network is presented with an entire message as a probe, then the problem boils down to deciding whether or not this message has been stored.
For this case, it has been shown that the missed detection rate is zero (i.e., messages which were previously stored are always recognized by the GBNN), and the false positive rate depends on the number of messages which have been stored in the network~\cite{gripon2011sparse}.

The retrieval rules studied in this paper (Section~\ref{sec:rules}) are specifically designed for the case where the probe contains missing values. GBNNs can also be used with inputs which contain errors (e.g., flipped bits), but the decoding rule must be changed significantly and the decoding rules studied in this paper are no longer applicable.

\subsection{Characteristics}
In a classic Willshaw network model, a unique activation threshold needs to be chosen globally, e.g., either the number of active neurons in the input probe or the maximum number of signals accumulated in output neurons.
This global threshold is no longer required in a GBNN, since each cluster can naturally decide for itself.
The most closely related model in the literature is that of Shim~\etal~\cite{shim1991statistical}.
However, there are two main advantages of GBNNs, both of which are supported by simulations in Section~\ref{sec:experiment}:
\begin{enumerate}
	\item GBNNs incorporate a self excitation term (see the reinforcement factor $\gamma$ in \eq{eq:oldindex:score} and \eq{eq:newrule:score}).
	\item The retrieval rule used in~\cite{shim1991statistical} is essentially the same as the \sos{} rule. Below we point out fundamental issues with \sos and we also advocate using the alternative \som{} rule~\cite{gripon2012nearly} for decoding, which significantly increases the retrieval rate.
\end{enumerate}

Sparseness has been heavily exploited to perform machine learning tasks and statistical inferences.
One of the most famous example is compressive sensing~\cite{baraniuk2007compressive}.
Neuroscientists are also aware of the sparseness principle~\cite{ganguli2012compressed}, not only because of its practical applications but also the low firing rate of the neurons in biological networks~\cite{golomb1990willshaw}.
The cluster structure of GBNN produces an extremely sparse binary code by definition (one active neuron per cluster), which makes the network biologically plausible, and also makes fast implementations possible.
It is also mentioned in~\cite{knoblauch2005neural} that the performance of an associative memory is severely affected given correlated patterns.
There, an intermediate ``grandmother cell'' layer is suggested to encode each pair of patterns using a neuron.
However, for GBNN, this particular problem can be mitigated by padding messages with extra random symbols at the cost of additional materials.
The cluster structure again makes the extension straightforward.

\section{Retrieval Rules\label{sec:rules}}
In this section, we review two existing retrieval rules for GBNN, i.e., \sos{} and \som{}.

\subsection{The \sos{} Rule}
The simplest rule~\cite{gripon2011sparse} is to add all the signals a neuron receives in the current iteration.
When presented with an incomplete message, we initialize the network by deactivating (i.e., setting to $0$) all the neurons within the clusters associated with erased symbols. We then repeat the following iterations. First, each neuron compute the sum of all connected neurons which are presently active. Then the neurons within each cluster with the most active connected neurons remain activated at the beginning of the next iteration.

Formally, let $\neuron(c,l)$ denote the $l$\textsuperscript{th} neuron in the $c$\textsuperscript{th} cluster, and let $w_{(cl)(c'l')}$ denote an indicator variable for whether or not a connection is present between $\neuron(c,l)$ and $\neuron(c',l')$; i.e.,
\begin{equation}
	\label{eq:oldindex:w}
	w_{(cl)(c'l')} =
	\begin{cases}
		1 & \mbox{$\neuron(c,l)$ connected to $\neuron(c',l')$}\\
		0 & \mbox{otherwise}
	\end{cases}.
\end{equation}
We also denote by $s_{cl}^t$ and $v_{cl}^{t}$ respectively the score function for the number of signals $\neuron(c,l)$ receives and the indicator function for whether or not $\neuron(c,l)$ is activated at iteration $t$, with $v_{cl}^0$ being the corresponding value for $\neuron(c,l)$ in the probe; i.e.,
\begin{equation}
	\label{eq:oldindex:v}
	v_{cl}^t =
	\begin{cases}
		1 & \quad\mbox{$\neuron(c,l)$ activated in iteration $t$}\\
		0 & \quad\mbox{otherwise}
	\end{cases}.
\end{equation} 
As a consequence, the retrieval procedure can be formalized as
\begin{align}
	s_{cl}^t & = \gamma v_{cl}^t + \sum_{c'=1}^{C}{\sum_{l'=1}^{L}{(v_{c'l'}^t w_{(c'l')(cl)})}}\label{eq:oldindex:score}\\
	s_{c,\max}^t & = \max_{1\leq l\leq L}{s_{cl}^t}\label{eq:findmax}\\
	v_{cl}^{t+1} & =
	\begin{cases}
		1 & \quad\mbox{if $s_{cl}^t = s_{c,\max}^t$}\label{eq:chooseMax}\\
		0 & \quad\mbox{otherwise}
	\end{cases},
\end{align}
where $\gamma \ge 0$ is a reinforcement factor.
Essentially, \eq{eq:oldindex:score} counts the score for each neuron.
It involves summing over all clusters and all neurons within each cluster, hence the name \sos{}.
\eq{eq:findmax} finds the value of the neurons with the strongest signal in each cluster, and \eq{eq:chooseMax} keeps them activated.

At the retrieval stage, the variables $w_{(cl)(c'l')}$ are fixed. These binary-valued variables are only changed when storing new messages.
The only parameter to be tuned for retrieval using \sos{} is $\gamma$, which influences the extent to which a neuron's own value influences its signal at the current iteration. 

\subsection{Problems with the \sos{} Rule}

The \sos{} rule, although straightforward and natural, might lead to unnecessary errors.
This is due to the fact that during iterations, after evaluating \eq{eq:chooseMax}, there might be multiple neurons in one cluster achieving the maximum value $s_{c,max}^t$ simultaneously.
In this case, all these neurons will stay activated and contribute to the signal strengths in the next iteration.

\begin{figure}
\centering
\includegraphics[scale=.5]{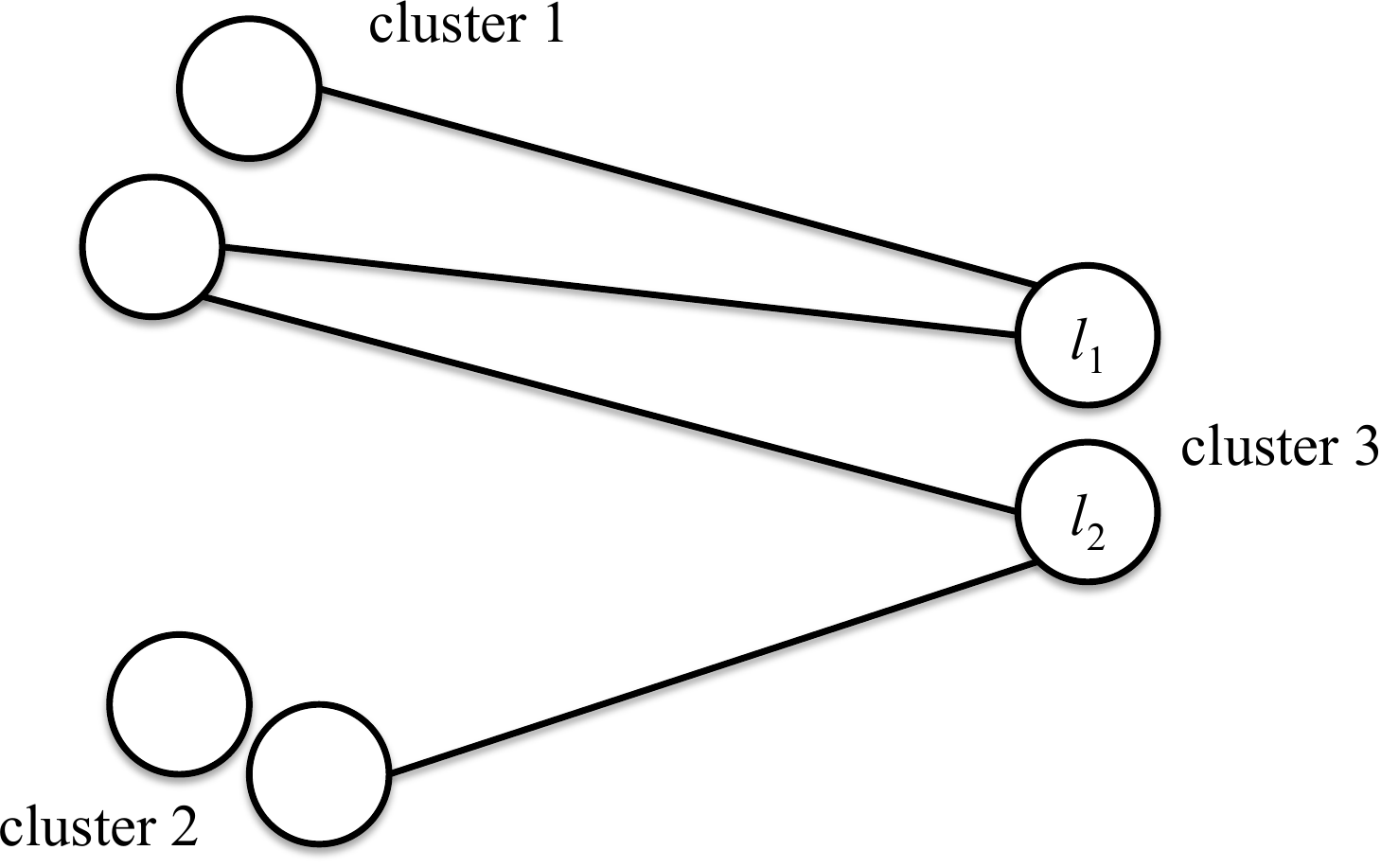}
\caption{Illustration of the \sos{} trap. Only the signals flowing into cluster $3$ are drawn.}
\label{fig:trap}
\end{figure}

Consider the scenario shown in \fig{fig:trap}, where two neurons $l_1$ and $l_2$ both receive the same number of signals.
Neuron $l_1$ receives two signals from cluster~$1$, while $l_2$ receives one signal from each cluster.
In this case, $l_2$ should be favored, because we know that for any individual pattern that has been stored, only one neuron in each cluster should be activated.
A possible but worse situation arises when $l_1$ receives more signals than $l_2$, since then $l_1$ will be the only activated neuron in this cluster at the beginning of the next iteration, even if $l_2$ was actually the correct neuron in cluster~$3$.
An increasing number of clusters will complicate the problem even further.
This can also cause \sos{} to diverge see Section~\ref{sec:property}.

\subsection{The \som{} Rule}

To avoid the problem mentioned in the previous subsection, the \som{} rule is proposed in~\cite{gripon2012nearly}. The rule is formally described as follows:
\begin{align}
	s_{cl}^t &= \gamma v_{cl}^t + \sum_{c'=1}^{C}{\max_{1\leq l'\leq L}{\left( v_{c'l'}^tw_{(c'l')(cl)}\right)}}\label{eq:newrule:score}\\
	v_{cl}^{t+1} &=
	\begin{cases}
		1 & \quad\mbox{if} \quad s_{cl}^t = \gamma + C - 1\\
		0 & \quad\mbox{otherwise}
	\end{cases}.\label{eq:newrule:select}
\end{align}
\eq{eq:newrule:score} involves a summation over \emph{max} operation, hence the name \som{}.
The basic idea is that, to retrieve the correct message, the score of a neuron should not be larger if it receives multiple signals from the same cluster, and the maximum taken in \eq{eq:newrule:score} ensures each neuron receives at most one signal from each cluster.
Since each stored message corresponds to a clique of $C$ neurons, one in each cluster, a neuron should be activated if it receives \emph{exactly} $C-1$ signals from the other clusters plus some value $\gamma$ from the self loop.

For \som{} to work properly, the network must be initialized appropriately when a probe is presented.
Instead of initializing all neurons associated with erased symbols to be $0$ as in \sos{}, we initialize them to be $1$. 
In that case, other neurons will definitely receive signals from these missing clusters, $L$ signals per missing cluster, but they will be regulated by \eq{eq:newrule:select}.

\section{Accelerating Retrieval\label{sec:acceleration}}

In this section, we will first briefly introduce the CUDA architecture.
We discuss different approaches to speeding up the GBNN retrieval procedure in general, and then we focus on specific techniques for \sos{} and \som{} separately.
We also illustrate graphically the dedicated CUDA kernel functions for both rules.
Note that, although we implement GBNN using CUDA, the accelerating techniques do not depend on any CUDA specific attribute, thus can be easily extended to other architectures.

\subsection{CUDA}
The \emph{Compute Unified Device Architecture} (CUDA), introduced in 2007, is NVIDIA's computing platform solution to \emph{general purpose computing on graphics processing units} (GPGPU), which enables dramatic increases in computing performance by harnessing the massively parallel resources of GPUs.
See~\cite{kirk2010programming} by Kirk and Hwu for more information.

\begin{figure}
\centering
\includegraphics[scale=.45]{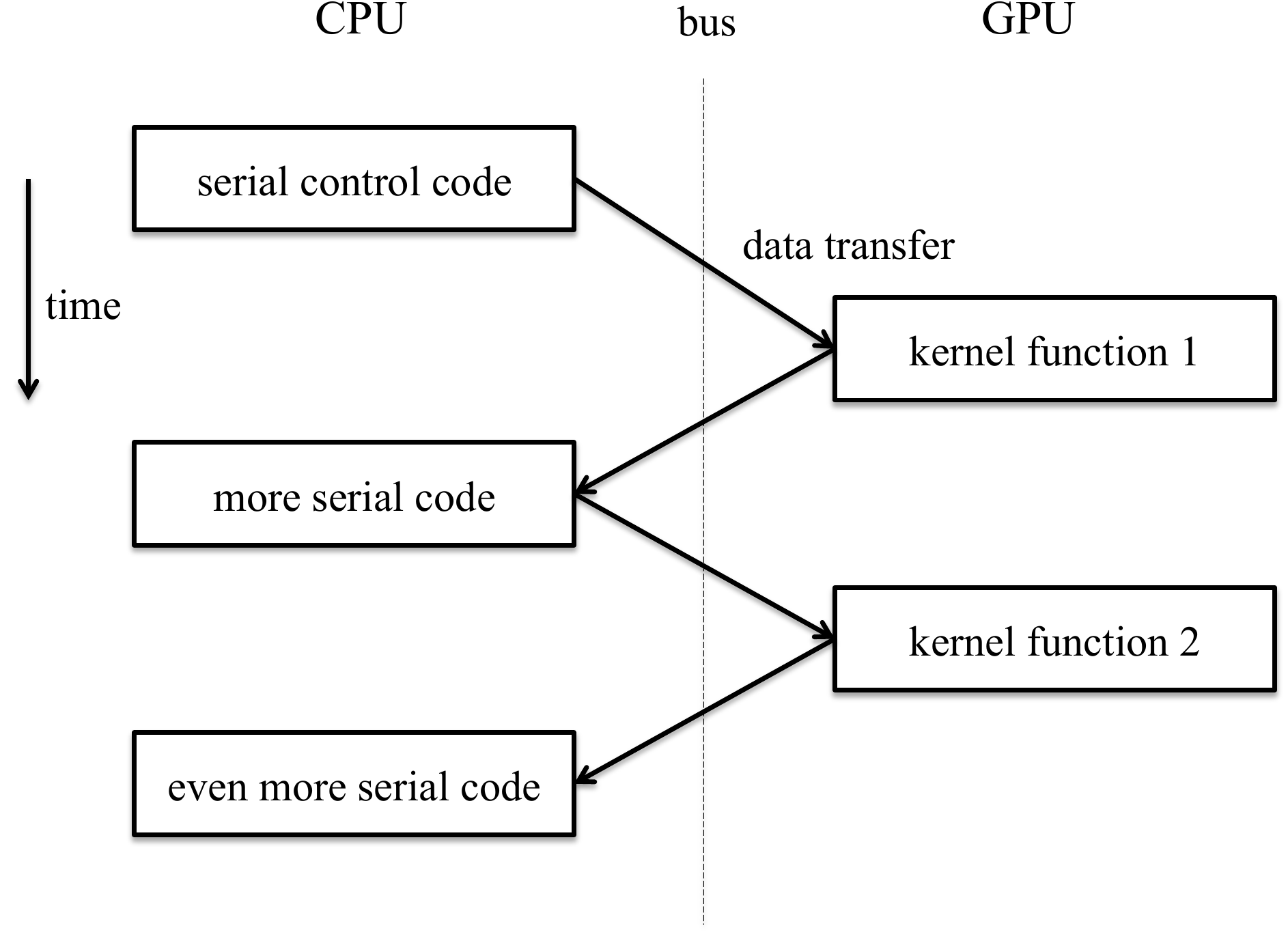}
\caption{CUDA programming scheme.}
\label{fig:cuda}
\end{figure}

The basic programming pattern in CUDA is as shown in~\fig{fig:cuda}, where CPUs play the role of managers, invoking on the GPUs some computational intensive functions called \emph{kernel} functions.
After the kernel function is executed on the GPU, the CPU collects the results back to the host and then may invoke more kernel functions if necessary.
Although a GPU can spawn many threads working simultaneously, each thread must run the same sequence of instructions. Kernel functions, and hence GPU computing in general, fit the category of ``single instruction multiple data'' (SIMD)~\cite{flynn1972some} parallel computing platforms.
The data are transferred back and forth between the CPU and GPU over the (slow) PCI or PCIe bus, one of the performance bottlenecks.
Unfortunately, since the code control flow is on the CPU side, the time-costly transfers between the host and the video card are inevitable.
Therefore, keeping the transfer of data to a minimum is one of the crucial concerns.

\subsection{General Tricks}
\subsubsection{Vectorization}
Although GBNN is a recurrent model, conceptually we can treat it as a layered network nevertheless.
We repeat each iteration $t$ as one layer, so that the number of layers can grow as large as the network needs to converge. Let $T$ denote the the total number of iterations to be run.
The only two constraints to be satisfied are 
\[
	w_{ij}^1 = w_{ij}^2 = \cdots = w_{ij}^T \quad\mbox{and}\quad w_{ij}^t = w_{ji}^t, t=1, 2, \cdots, T.
\]

The benefit is to borrow the matrix notation from layered networks which is more efficient in the parallel implementation.
We map the original clustered structure into a flat space, where $\neuron(c,l)$ becomes $\neuron(i)$, with $i = (c-1)L+l$, ranging from $1$ to $n = CL$.
Then \eq{eq:oldindex:w} and \eq{eq:oldindex:v} can be rewritten as
\begin{align}
	w_{ij} &=
	\begin{cases}
		1 & \quad\mbox{$\neuron(i)$ connects $\neuron(j)$}\\
		0 & \quad\mbox{otherwise}
	\end{cases}\\
	v_{i}^t &=
	\begin{cases}
		1 & \quad\mbox{$\neuron(i)$ is activated in iteration $t$}\\
		0 & \quad\mbox{otherwise}
	\end{cases}.
\end{align}

We consider the edge weights $w_{ij}$ as elements of an $n\times n$ matrix $W$ and neuron potentials $v_i$ as elements of a vector $v \in \{0,1\}^n$.
Taking into account the reinforcement factor $\gamma$, we can rewrite \eq{eq:oldindex:score} as
\begin{equation}
	s^t = Wv^t\label{eq:serial},
\end{equation}
with $W$ being a symmetric matrix whose diagonal elements are all equal to $\gamma$ and whose off-diagonal elements are all binary valued; i.e.,
\[W = \begin{pmatrix}
	\gamma & w_{12} & \cdots & w_{1n}\\
	w_{21} & \gamma & \cdots & w_{2n}\\
	\vdots & \vdots & \ddots & \vdots\\
	w_{n1} & w_{n2} & \cdots & \gamma 
\end{pmatrix}.\]
Thus, the score equation~\eqref{eq:serial} is a matrix-vector product, computed efficiently in parallel on a GPU.

\subsubsection{Batch Retrieval}
A straightforward extension to vectorization is to bundle and process $K$ probes simultaneously. To do so, we collect the $K$ test messages into a value matrix
\[
	V^t = \begin{pmatrix}
		v^t(1) & v^t(2) & \cdots & v^t(K)
	\end{pmatrix},
\]
with each column $v^t(k)$ being a value vector in \eq{eq:serial}, so that \eq{eq:serial} becomes
\begin{equation}
	S^t = WV^t.\label{eq:parallel}
\end{equation}
Instead of retrieving messages one after another, we aggregate $K$ messages together and feed them into the GPU card at one shot.
Speedups are achieved using this approach because it allows us to exploit the SIMD nature of GPUs.
It is also more efficient to perform one large I/O transfer over the bus rather than multiple smaller transfers.

Batch retrieval arises naturally in applications where simultaneous retrievals are preferred.
For instance, in face recognition, an associative memory can be used to recognize face features even when areas are obstructed by sun glasses or a scarf.
If we treat each image as a single message, the hardware requirement is simply prohibitive.
A $256$-level gray image of the size $512\times 512$ requires an adjacency matrix $W$ of $(2^{9}\times 2^{9}\times 2^{8})^2 = 2^{52}$ elements.
Alternatively, we can divide the image into smaller patches, treat each patch as a different message, and process them in parallel.
For another example, consider a network anomaly detection algorithm where we are given a batch of IP addresses, and we would like to check whether each belongs to a predefined blacklist.
In Section~\ref{sec:experiment} below, we will refer to \eq{eq:parallel} as parallel decoding and \eq{eq:serial} as serial decoding.

\subsubsection{Reduction}
\begin{figure}
\centering
\includegraphics[scale=.5]{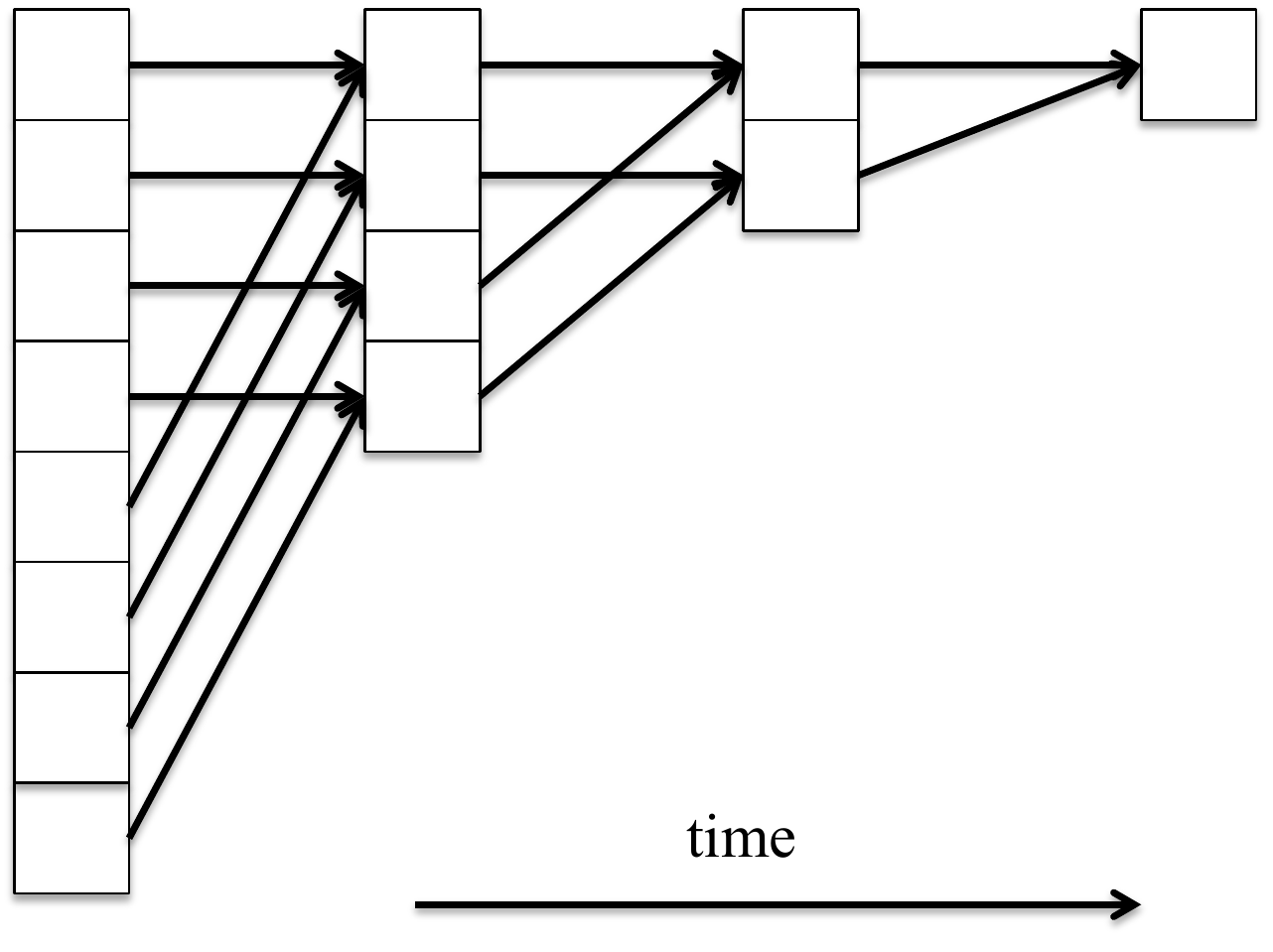}
\caption{Parallel reduction operation scheme.}
\label{fig:reduction}
\end{figure}

Reduction refers to an operation that aggregates a vector of elements into a scalar (e.g., \emph{sum}, \emph{max} and \emph{min}).
In \sos{}, the \emph{max} operation is needed when evaluating \eq{eq:findmax} to determine which neurons remain active in the next iteration.
In both rules, when deciding whether or not the retrieval procedure has converged, we need to compare two long vectors $v^t$ and $v^{t+1}$ of length $n$, and test if all of the neuron values stay unchanged.
This reduction operation can be done in an efficient manner as illustrated in \fig{fig:reduction}, where we invoke $n$ threads in the first step, afterwards halving the number of threads in every successive operation.
The time complexity thus decreases from $O(n)$ to $O(\log_2{n})$.

\subsubsection{Sparsification}
Memory access is an expensive operation on the video card, where both reading from and writing to memory cells are much slower than on the host CPU.
In order to combat this inefficiency, we can reduce the number of memory accesses by accounting for the fact that GBNN is actually a sparse network; i.e., for a given message, ideally only one neuron should be activated for each cluster.
Typically, the network structure should also be sparse, so we could implement both $W$ and $V^t$ as sparse matrices using compressed format, where we only record the nonzero elements and their coordinates. Then evaluating \eq{eq:oldindex:score} and \eq{eq:newrule:score} requires many fewer terms.
However, the compressed format does not lead to a significant performance gain for both rules --- \som{} benefits from the sparsification, while \sos{} does not.
The reason is that the dense matrix product in \eq{eq:parallel} for \sos{} is an optimized operation on the GPU, whereas the compressed format deviates from the optimized pattern.
Moreover, since $V^t$ changes from one iteration to the next, it is not economical to implement $V^t$ using compressed format either.
On the contrary, $W$ is fixed at the retrieval stage.
We use a sparse matrix representation $W$ only for \som{}.
Detailed numerical results are presented in Section \ref{sec:experiment}.

\subsection{Accelerating the \sos{} Rule}

The pseudocode for the \sos{} procedure is given in Algorithm \ref{alg:sumofsum}.
It requires as inputs the maximum number of iterations permitted $t_{max}$, the weight matrix $W$ with all of the clique structures preserved during the storing stage, and the message matrix $V^0$, with the $k$\textsuperscript{th} column being the value vector for test message $k$ and the erased clusters deactivated.
On Line~\ref{line4}, $S^t$ is the score matrix for iteration $t$, where the $k$\textsuperscript{th} column is the score vector of length $n$ for test message $k$.
On Line~\ref{line5}, the kernel function takes $S^t$ as input and essentially produces $V^{t+1}$ by evaluating \eq{eq:findmax}.

\begin{algorithm}
\caption{The \sos{} retrieval procedure.}
\label{alg:sumofsum}
\begin{algorithmic}[1]
\REQUIRE    the maximum number of iterations permitted $t_{max}$, the weight matrix $W$, the message matrix $V^0$ with each column as a partially erased message for recovery
\ENSURE     the recovered matrix $V^t$
\\~\\
\STATE		t = -1
\REPEAT  
\STATE		t = t+1
\STATE 	$S^t = WV^t$\label{line4}
\STATE		$V^{t+1} = \mbox{the kernel function as in \fig{fig:sumofsum}}$\label{line5}
\UNTIL		{$V^{t+1} == V^{t}$ or $t == t_{max}$}
\end{algorithmic}
\end{algorithm}

\begin{figure}
\centering
\includegraphics[scale=.6]{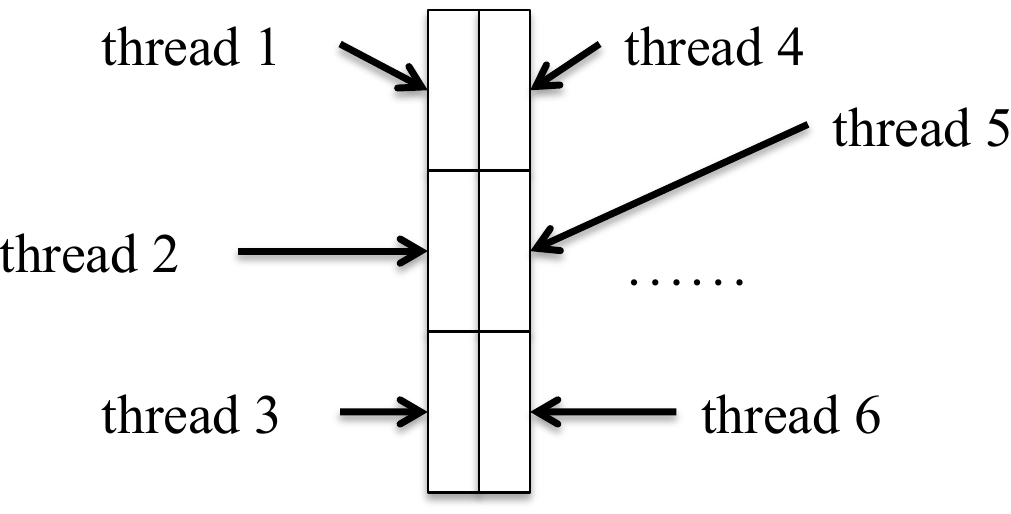}
\caption{Illustration of the kernel function for the \sos{} rule. The first two columns of $S^t$ are drawn. Each rectangular represents a cluster with $L$ neurons. A thread will determine the maximum value in its cluster and set the corresponding neurons activated.}
\label{fig:sumofsum}
\end{figure}

The first two columns of $S^t$ are drawn in \fig{fig:sumofsum}.
In this particular example, each message can be divided into $C=3$ clusters.
In our implementation, a dedicated thread processes one cluster, finding the maximum value in that cluster, and then keeping the neurons that reach the maximum value activated.
Assuming that there are $K$ messages to be recovered, a total of $CK$ threads are used.
The retrieval procedure terminates when either the network converges or it reaches the maximum number of iterations permitted.
 
\subsection{Accelerating the \som{} Rule}

\begin{figure}
\centering
\includegraphics[scale=.6]{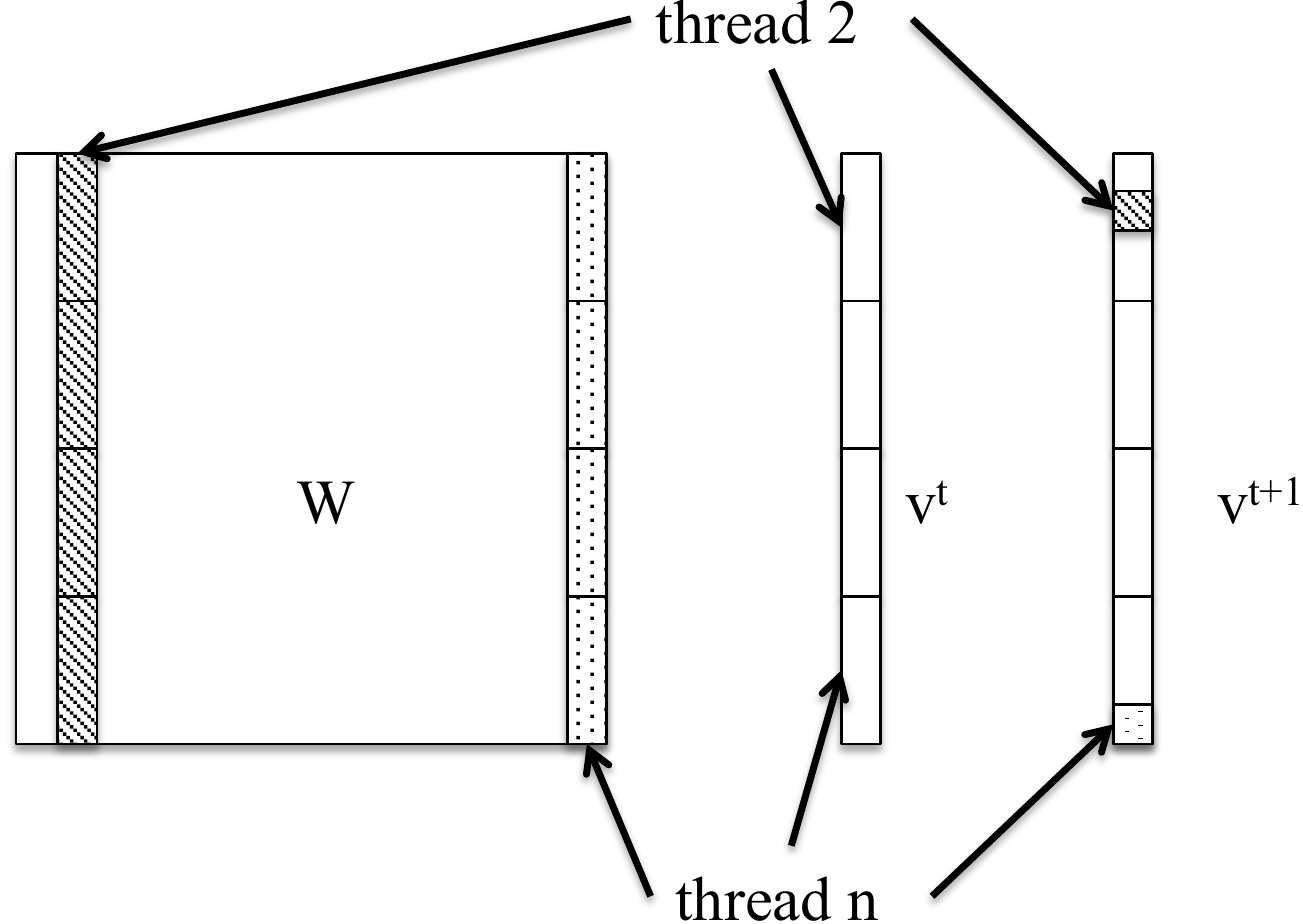}
\caption{Illustration of the kernel function for the \som{} rule. A single message retrieval is shown. To update the element $v^{t+1}_{i}$, we examine both $v^t$ and the $i$\textsuperscript{th} column of $W$.}
\label{fig:sumofmax}
\end{figure}

The pseudocode for \som{} is almost the same with Algorithm~\ref{alg:sumofsum}, except that Lines~\ref{line4} and \ref{line5} are replaced by another kernel function illustrated in \fig{fig:sumofmax}.
In order to better explain the concept, the serial decoding of a single message $v^t$ is presented here, where the same number of threads are needed as the number of neurons $n$ in the network.
The extension to the parallel decoding scheme of $K$ bundled messages is straightforward, where $nK$ threads are needed.

We do not follow strictly \eq{eq:newrule:score} and \eq{eq:newrule:select} to evaluate a \emph{max} function.
Instead, we apply an alternative procedure.
Essentially, we check if a neuron receives signals from every cluster; hence, for $\neuron(i)$, the $i$\textsuperscript{th} row of $W$ and $v^t$ require examination.
Since $W$ is symmetric and the memory storage on the GPU is column major, we check the $i$\textsuperscript{th} column of $W$ instead to make the computation more pleasant.
To update a neuron value $v_{i}^{t+1}$, a dedicated thread $i$ is required, scanning through both $v^{t}$ and the $i$\textsuperscript{th} column of $W$.

Thread $i$ loops through cluster $c$, from $1$ to $C$.
\begin{itemize}
\item For any positive $\gamma$, if $\neuron(i)$ belongs to cluster $c=\lfloor\frac{i-1}{L}\rfloor+1$, we directly set $s_i^t=s_i^t+1$. ($\lfloor\cdot\rfloor$ is the standard floor operator.)
\item Otherwise, we check within the same cluster, i.e., $w_{ji}$ and $v_j^t$, where $j$ goes from $(c-1)L+1$ to $cL$.
The first time we encounter $w_{ji}>0$ and $v_j^t>0$, we set $s_i^t=s_i^t+1$, and proceed to the next cluster $c+1$ without further investigation.
\item If cluster $c$ does not contribute any signal to $\neuron(i)$, i.e., $s_i^t$ does not change, we stop right away without checking following clusters.
\end{itemize}

We call this procedure \boe{} and favor it over \eq{eq:newrule:score} and \eq{eq:newrule:select} for two reasons: 
\begin{enumerate}
\item It explicitly clarifies the requirement that every cluster should contribute one and only one signal.
\item It proceeds to subsequent clusters or stops processing as quickly as possible so that further expensive memory accesses are avoided.
\end{enumerate}

\begin{theorem}
\label{thm:bailout}
    The \boe{} approach is equivalent to \som{}, i.e., for any positive $\gamma$, given $W$ and $v^t$, \boe{} produces the same $v^{t+1}$ as \som{}.
\end{theorem}
\begin{IEEEproof}
For cluster $c=\lfloor\frac{i-1}{L}\rfloor+1$, there is only one nonnegative $w_{ii}=\gamma>0$, since by design, within the same cluster, a neuron can only receive contributions from itself.
\boe{} directly sets $s_i^t=s_i^t+1$, which effectively makes any positive $\gamma$ equivalent to $\gamma=1$.
 
For other clusters, $w_{ji}$ is either $0$ or $1$, depending on whether or not a connection exists between $\neuron(j)$ and $\neuron(i)$.
Notice in either case, $v^t$ is always a binary vector.

Consider $\{a_j=v^t_jw_{ji}\}$, for $j=1,2,\cdots,L$.
The \boe{} approach treats \eq{eq:newrule:score} recursively by implementing the following equation,
\begin{equation}
	\max(a_1,a_2,\cdots,a_L)=\begin{cases}
		a_1 & \quad\mbox{if $a_1>0$}\\
		\max(a_2,\cdots,a_L) & \quad\mbox{otherwise}
	\end{cases} \ ,
\end{equation}
and treats \eq{eq:newrule:select} by setting $\gamma=1$.
\end{IEEEproof}

For a neuron to be activated, it needs to receive $C-1$ signals from the other clusters, plus some self contribution $\gamma$ from itself.
Since any $\gamma > 0$ does not affect the dynamic due to \eq{eq:newrule:score} and \eq{eq:newrule:select}, we deliberately set $\gamma=1$.
The weight matrix $W$ thus becomes binary valued, which can be more efficiently implemented.

\section{Properties\label{sec:property}}
In this section, we discuss properties of GBNNs and the two retrieval rules introduced in previous sections. We illustrate these properties via examples and theoretical claims.

\subsection{The \sos{} Rule May Oscillate}
We first give an example which illustrates that \sos{} may oscillate.
Consider a small network with $C=3$ clusters, each cluster has $L=3$ neurons, i.e., $9$ neurons in total.
We set $\gamma=1$.
There are $4$ messages to store:
$(1, 1, 1)$, $(2, 2, 1)$, $(3, 2, 1)$ and $(1, 3, 1)$.
The test message is $(?, ?, 1)$. Clearly all of the stored messages match the non-erased part of the test message. In such a scenario, we expect that the retrieval rule either returns an \emph{arbitrary} stored message which matches the input, or returns \emph{all} of the stored messages matching the input. Unfortunately, \sos{} does not converge to any output. After constructing the network and initializing the neurons to be deactivated for the 1\textsuperscript{st} and 2\textsuperscript{nd} clusters of the test message, we have
\[
W =\begin{pmatrix}
	1 & 0 & 0 & 1 & 0 & 1 & 1 & 0 & 0\\
	0 & 1 & 0 & 0 & 1 & 0 & 1 & 0 & 0\\
	0 & 0 & 1 & 0 & 1 & 0 & 1 & 0 & 0\\
	1 & 0 & 0 & 1 & 0 & 0 & 1 & 0 & 0\\
	0 & 1 & 1 & 0 & 1 & 0 & 1 & 0 & 0\\
	1 & 0 & 0 & 0 & 0 & 1 & 1 & 0 & 0\\
	1 & 1 & 1 & 1 & 1 & 1 & 1 & 0 & 0\\
	0 & 0 & 0 & 0 & 0 & 0 & 0 & 1 & 0\\
	0 & 0 & 0 & 0 & 0 & 0 & 0 & 0 & 1
\end{pmatrix}
\quad\mbox{and}\quad
v^0 = \begin{pmatrix}
	0\\0\\0\\0\\0\\0\\1\\0\\0
\end{pmatrix}.
\]
It is easy to verify that
\begin{align*}
s^0 &= (\underline{1},\underline{1},\underline{1},\underline{1},\underline{1},\underline{1},\underline{1},0,0)^T, & v^1 &= (1,1,1,1,1,1,1,0,0)^T,\\
s^1 &= (\underline{4},3,3,3,\underline{4},3,\underline{7},0,0)^T, & v^2 &= (1,0,0,0,1,0,1,0,0)^T,\\
s^2 &= (\underline{2},\underline{2},\underline{2},\underline{2},\underline{2},\underline{2},\underline{3},0,0)^T, & v^3 &= (1,1,1,1,1,1,1,0,0)^T.
\end{align*}
The underlined values indicate the maximum value within the same cluster.
Note that every cluster activates its own neurons with the most signals respectively.
Therefore, three neurons are activated in $v^2$.
In this case, $v^3 = v^1$, so that the network does not converge, oscillating between $v^2$ and $v^3$ forever.

There is another level of complication: the reinforcement factor $\gamma$ plays a delicate role in the retrieval procedure.
If we increase $\gamma = 2$, then the network converges. However, we will see in Section \ref{sec:experiment} below that enlarging $\gamma$ leads to a worse retrieval rate in general.

\subsection{The \som{} Rule Converges}
We now show that \som{} (\boe{}) always converges when all the neurons in erased clusters are initialized to be activated.

\begin{lemma}
\label{thm:dead}
	For \som{}, once deactivated, a neuron stays deactivated forever, i.e., if $v_i^t=0$ then $v_i^{t+1}=0$.
\end{lemma}
\begin{IEEEproof}
	Recall, from \eq{eq:newrule:select}, that $v_i^{t+1} = 1$ if and only if $s_i^t = \gamma+C-1$.
	Assume in iteration $t$ that $\neuron(i)$ is deactivated, i.e., $v^t_i = 0$. Then $v^t_iw_{ii} = 0$.
	Since the only possible contribution a neuron might obtain from its own cluster is the self loop, $s^t_i =\sum_{c=1}^{C}{\max_{j}(v^t_jw_{ji})}<\gamma+C-1$, thus $v^{t+1}_i = 0$.
\end{IEEEproof}

\begin{lemma}
\label{thm:stable}
	A clique is stable, i.e., once all neurons of a clique are activated, they stay activated forever.
\end{lemma}
\begin{IEEEproof}
	By definition of a complete sub-graph, all neurons in an activated clique (see \fig{fig:network}) will receive exactly $C-1$ signals from other clusters and some positive feedback $\gamma$.
	Therefore by \eq{eq:newrule:select}, all neurons in the clique stay activated in the next iteration.
\end{IEEEproof}

\begin{lemma}
\label{thm:ensemble}
	Given a partially erased message, \som{} always converges to a state which contains an ensemble of cliques.
\end{lemma}
\begin{IEEEproof}
	As each previously stored message corresponds to a clique, a partially erased message corresponds to parts of the clique, with the neurons in the non-erased clusters activated.
	\som{} initializes all the neurons in the missing clusters to be activated.
	Therefore, the already activated neurons in non-erased clusters will receive contributions from the missing clusters, staying activated in the next iteration.
	The neurons in the missing clusters which, together with the already activated neurons in non-erased clusters, form a clique will also receive exactly $\gamma+C-1$ signals and will stay activated in the next iteration.
	By Lemma~\ref{thm:stable}, the ensemble of these cliques will be present in the converged state.
\end{IEEEproof}

\begin{theorem}
\label{thm:live}
	Given a partially erased message, if a neuron is the only one activated in its cluster, it remains activated, i.e., for a given cluster $c$, if there exists an $i \in \{(c-1)L+1, \cdots, cL\}$ such that $v_i^t = 1$ and $v_j^t = 0$ for all $j \in \{(c-1)L+1, \cdots, cL\}$, $j \ne i$, 
then $v_i^{t+1}=1$.
\end{theorem}
\begin{IEEEproof}
	Suppose, to arrive at a contradiction, that at some point, cluster $c$ has no neuron activated, i.e., $\forall i=(c-1)L+1, \cdots, cL, v_i^t=0$, the other clusters will not receive any signal from cluster $c$.
	By \eq{eq:newrule:select}, every neuron throughout the network will be deactivated in the next iteration.
	By Lemma~\ref{thm:dead}, the network converges to this all-deactivated state forever, which violates Lemma~\ref{thm:ensemble}.
	Therefore, if a neuron is the only one activated in its cluster, it remains activated.
\end{IEEEproof}

\begin{theorem}
\label{thm:converge}
	For any given probe pattern, \som{} always converges.
\end{theorem}
\begin{IEEEproof}
	For a partially erased message, this theorem has already been proved by Lemma~\ref{thm:ensemble}.

	We consider an input probe such that some parts of a previously stored message are modified (corrupted).
	If the probe can still be explained by a clique in the network, the memory converges to this clique by Lemma~\ref{thm:stable}.
	If the probe cannot be explained by any clique in the network, the activated neurons in the unchanged clusters cannot receive signals from the corrupted clusters.
	Hence by \eq{eq:newrule:select}, the memory converges to the all-deactivated state.
\end{IEEEproof}

Since \boe{} is equivalent to \som{} (see Theorem~\ref{thm:bailout}), we also have the following.
\begin{corollary}
For any given probe pattern, \boe{} always converges.
\end{corollary}

It is worth emphasizing that \som{} converges to a state which contains an ensemble of cliques by Lemma~\ref{thm:ensemble}.
We can randomly choose one of them as the reconstructed message.

\section{Joint Retrieval Rule\label{sec:joint}}
\subsection{Proposal}
We have just seen that \sos{} is not guaranteed to converge, whereas \som{} is.
In Section~\ref{sec:experiment} below we will see that \sos{} is generally much faster than \som{}, but the accuracy of \som{} is much better when either the number of stored messages or the number of erased symbols increases.
It is natural to ask whether we can merge the two rules to obtain a fast and accurate reconstruction scheme.

In this section we propose such a hybrid retrieval scheme which combines the best aspects from both procedures.
The pseudocode for the joint decoding scheme is given in Algorithm~\ref{alg:joint}.
Essentially this decoding algorithm performs one refined iteration of \sos{} followed by subsequent, optimized iterations of \boe{} until a convergence criterion is satisfied.

\begin{algorithm}
\caption{The joint retrieval scheme.}
\label{alg:joint}
\begin{algorithmic}[1]
\REQUIRE    
$C$ -- number of clusters\\
$L$ -- number of neurons in each cluster\\
$e$ -- number of erased clusters\\ 
$K$ -- number of test messages\\
$W$ -- the weight matrix of dimension $n\times n$\\
$V^0$ --  the value matrix of dimension $n\times K$ with each column as a partially erased message for recovery
\ENSURE     the recovered matrix $V^t$
\\~\\
\STATE		initialize all the neurons inactive in erased clusters
\STATE		$S^0 = WV^0$ \COMMENT{\sos{}}
\FOR[$eK$ threads in parallel, each for an erased cluster]{each thread}
	\STATE check $S^0$ and keep neurons with $C-e$ signals activated
\ENDFOR \COMMENT{$V^1$ obtained}
\STATE		sparsify $W$ to obtain $W'$ \COMMENT{use $W'$ afterwards}
\STATE		t = 0
\REPEAT[\boe{} (\som{})]
	\STATE		t = t+1
	\FOR[$eLK$ threads in parallel, each for a neuron in erased clusters of different messages]{each thread}
			\STATE keep off deactivated neurons, otherwise apply the \boe{} scheme
	\ENDFOR\quad\COMMENT{$V^{t+1}$ obtained}
\UNTIL		{$V^{t+1} == V^{t}$}\quad\COMMENT{only check erased clusters}
\end{algorithmic}
\end{algorithm}

\subsection{Justification and Convergence}

As mentioned in Section~\ref{sec:acceleration}, memory access is extremely expensive on GPUs in comparison to on the host CPU. Therefore, it is of vital importance that we eliminate any unnecessary memory operations.
We notice that Lemma~\ref{thm:dead} and Theorem~\ref{thm:live} have crucial implications in designing our new scheme.
The former suggests that if $v^t_i=0$ then there is no need to loop through two long vectors of length $n$, i.e., $v^t$ and the $i$\textsuperscript{th} column of $W$, since we will have $v_i^{t+1}=0$.
Thus, we only need to focus on updating those $\neuron(i)$ for which $v_i^t = 1$. In this sense, the currently active neurons can be considered as a candidate pool that needs to be investigated further in the next iteration.
The latter suggests that clusters with only one active neuron (including those which are not erased in the test message) will not change during decoding. Hence, we only update neurons in erased clusters that have not yet reached convergence. In general, this notion of ``freezing good clusters'' can also be justified as preventing good neurons from being altered undesirably by any retrieval procedure.

One final but important consideration is the all-activated initialization scheme.
Although it is crucial for the correctness of the \som{} rule, it also introduces too many candidates from the beginning.
We will show a motivating example later in Section~\ref{motivation}.
Fortunately, \sos{} can help us bypass this particular problem.

\begin{theorem}
\label{thm:joint}
	The first iteration of \sos{} affects neither the correctness of \som{} nor its convergence.
\end{theorem}
\begin{IEEEproof}
For correctness, let us revisit \sos{}.
The only problem making \sos{} inferior to \som{} is that during the retrieval procedure, as in \fig{fig:trap}, it is possible for multiple neurons to be activated simultaneously in one cluster \emph{without} regulation, which in turn propagates the error or even causes oscillation.
However, during the initialization phase, if we deactivate all the neurons in erased clusters, preserving good clusters only, by definition there will be at most one activated neuron per cluster.
The aforementioned flaw does not exist anymore.

For convergence, recall a clique structure in \fig{fig:network}.
For a given message with $e$ clusters erased, there are $C-e$ good neurons transmitting signals.
Therefore, the desired neuron in the erased clusters should receive $C-e$ signals exactly.
After one iteration of \sos{}, we only keep these neurons with $C-e$ signals in the candidate pool.
The sole effect of the first iteration \sos{} is to shrink the pool size, with the convergence untouched.
\end{IEEEproof}

Ideally, there should be only one such neuron per erased cluster in the candidate pool, rather than $L$ candidates for \som{}, with two exceptional cases.
\begin{IEEEitemize}
	\item There are two memorized messages $m_1$ and $m_2$ which only differ in erased clusters, e.g., we have $m_1=(1,3,1)$ and $m_2=(1,3,2)$, and the test message is $m=(1,3,?)$.
	In this case, both neuron $1$ and $2$ in the erased cluster will be present in the pool.
	\item Spurious cliques. While storing messages, distinct cliques may overlap and produce a spurious clique that does not correspond to a stored message, where different edges in this clique were added for different stored messages. In other words, a stored message corresponds to a clique, but not vice versa.
\end{IEEEitemize}
We argue that for a relatively large network and a reasonable number of messages to memorize, the candidate pool size is sufficiently small.

\begin{corollary}
	The new joint retrieval scheme always converges.
\end{corollary}
\begin{IEEEproof}
	The joint scheme invokes one iteration of \sos{} followed by iterations of \boe{}.
	According to Theorem \ref{thm:joint}, \sos{} effectively reduces the size of the candidate pool, with the network's convergence property untouched.
	The joint scheme thus always converges by Theorem \ref{thm:converge}.
\end{IEEEproof}

Combining all these factors, we propose the joint scheme as in Algorithm \ref{alg:joint}.

\section{Experiments\label{sec:experiment}}

In this section, we compare \sos{} and \som{} using the different acceleration approaches discussed previously in Section~\ref{sec:acceleration} and Section~\ref{sec:joint}.
We show that a significant performance gain is achieved in terms of running time after parallelizing the retrieval procedure and applying our new joint retrieval scheme.

All the CPU experiments are executed on a $2.6$GHz AMD Phenom (tm) 9950 Quad-Core Processor with $16$GB of RAM, and all the GPU experiments are executed on an NVIDIA C1060 card, which runs at a frequency of $1.3$GHz with $4$GB memory and has $30$ stream multiprocessors.
In order to make as fair a comparison as possible, our CPU code makes use of the Armadillo library~\cite{sanderson2010armadillo}, linked to BLAS and LAPACK, for optimized linear algebra operations.

\subsection{\sos{} versus \som{}}

\begin{figure}
\centering
\includegraphics[scale=.5]{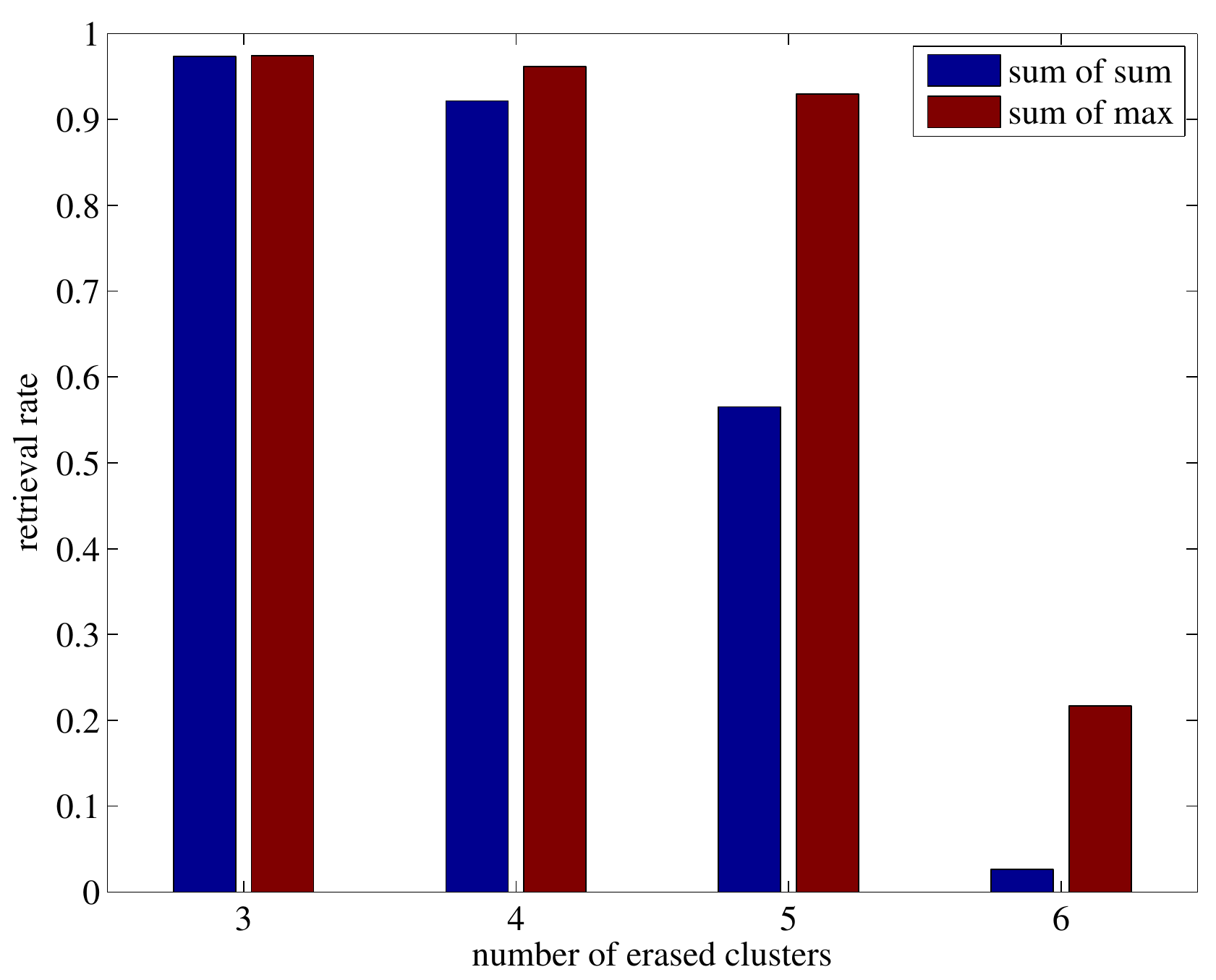}
\caption{Comparison between the \sos{} and \som{} retrieval rules. We have $C=8$, $L=128$, $5000$ stored messages and $3000$ test messages. We vary the number of erased clusters, and plot the retrieval rate.}
\label{fig:cmprules}
\end{figure}

First, we compare \sos{} and \som{}.
In this experiment, we have $C=8$ clusters with $L=128$ neurons each, and the reinforcement factor $\gamma=2$.
We randomly generate and store $5000$ messages, each of which consists of $8$ symbols uniformly sampled from the integers $1$ to $128$.
After the storing stage, we randomly select $3000$ out of the $5000$ stored messages, erase some parts of them and try to retrieve the erased messages from the GBNN associative memory.
We refer to this experiment setting as Scenario~$1$.
Since \sos{} does not necessarily converge, we set the maximum number of iterations to $20$.
We vary the number of erased clusters, and plot the retrieval rate, i.e., the fraction of successfully retrieved messages, in \fig{fig:cmprules}.

Observe that when the number of erased clusters is relatively small ($3$ erased clusters), both rules perform equally well above $97\%$.
As the number of erased clusters increases, although both rules make more errors, the performance of \sos{} degrades more drastically than that of \som{}.
When $5$ out of $8$ clusters are erased, \sos{} can only recover slightly above $50\%$ of the messages, while \som{} still recovers over $90\%$.
If $6$ clusters are erased, \som{} is still able to retrieve over $20\%$, which is significantly higher than \sos{}. 
\subsection{Influence of \texorpdfstring{$\gamma$}{gamma}}

\begin{figure}
\centering
\includegraphics[scale=.5]{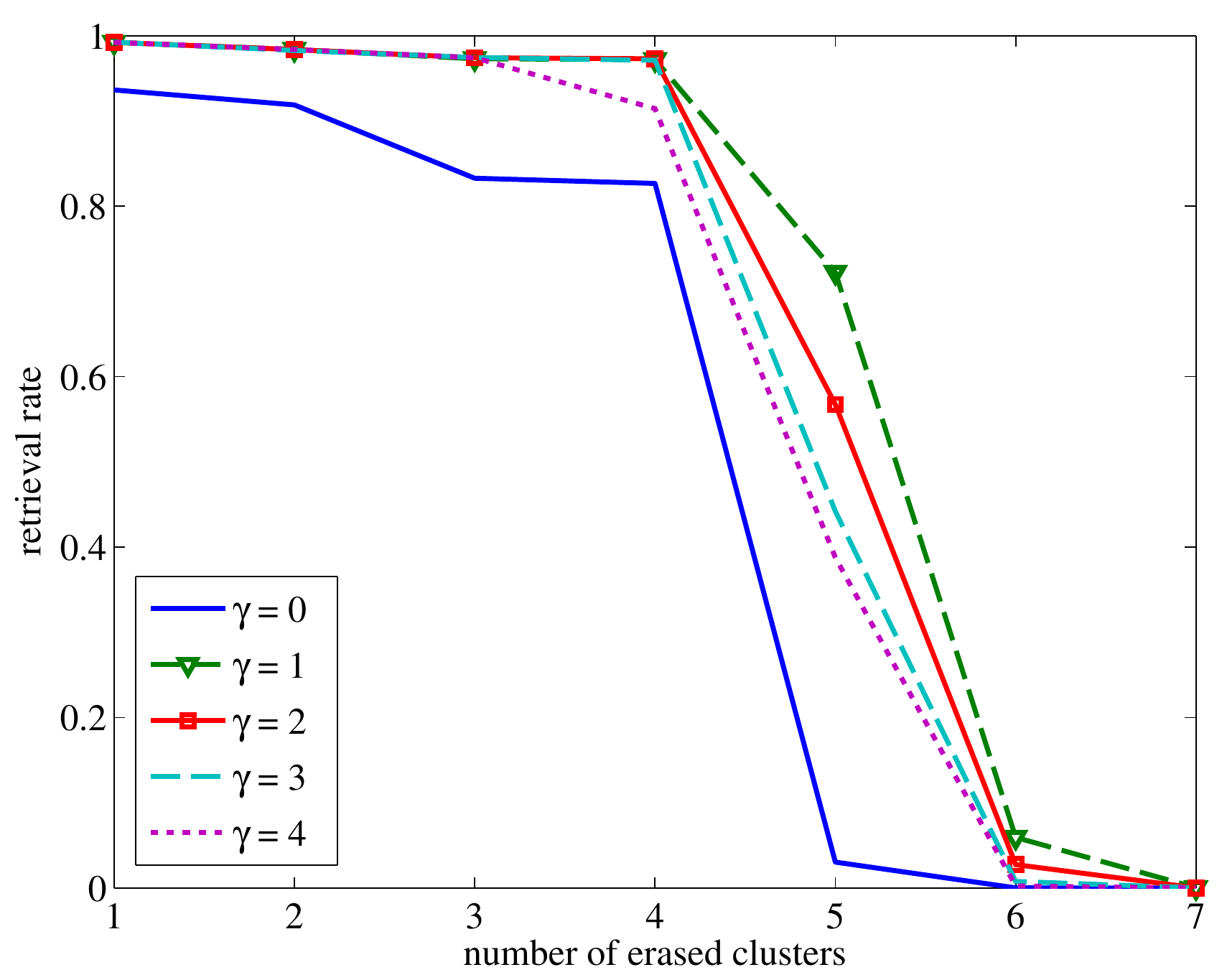}
\caption{Subtle dependence between the retrieval rate and the reinforcement factor $\gamma$. We have $C=8$, $L=128$, $5000$ stored messages and $3000$ test messages. We vary the number of erased clusters, and plot the retrieval rate.}
\label{fig:gamma}
\end{figure}

Second, we explore the subtle dependence of the retrieval rate on the value of the reinforcement factor $\gamma$ used in \sos{}.
We plot the trend for different $\gamma$ in \fig{fig:gamma} using the same experiment Scenario~$1$ as above.
In general, increasing $\gamma$ hurts the retrieval rate with the only exception of $\gamma=0$, which suggests that $\gamma = 1$ can be used as a default value.

\subsection{CPU versus GPU}

\begin{figure}
\centering
\includegraphics[scale=.5]{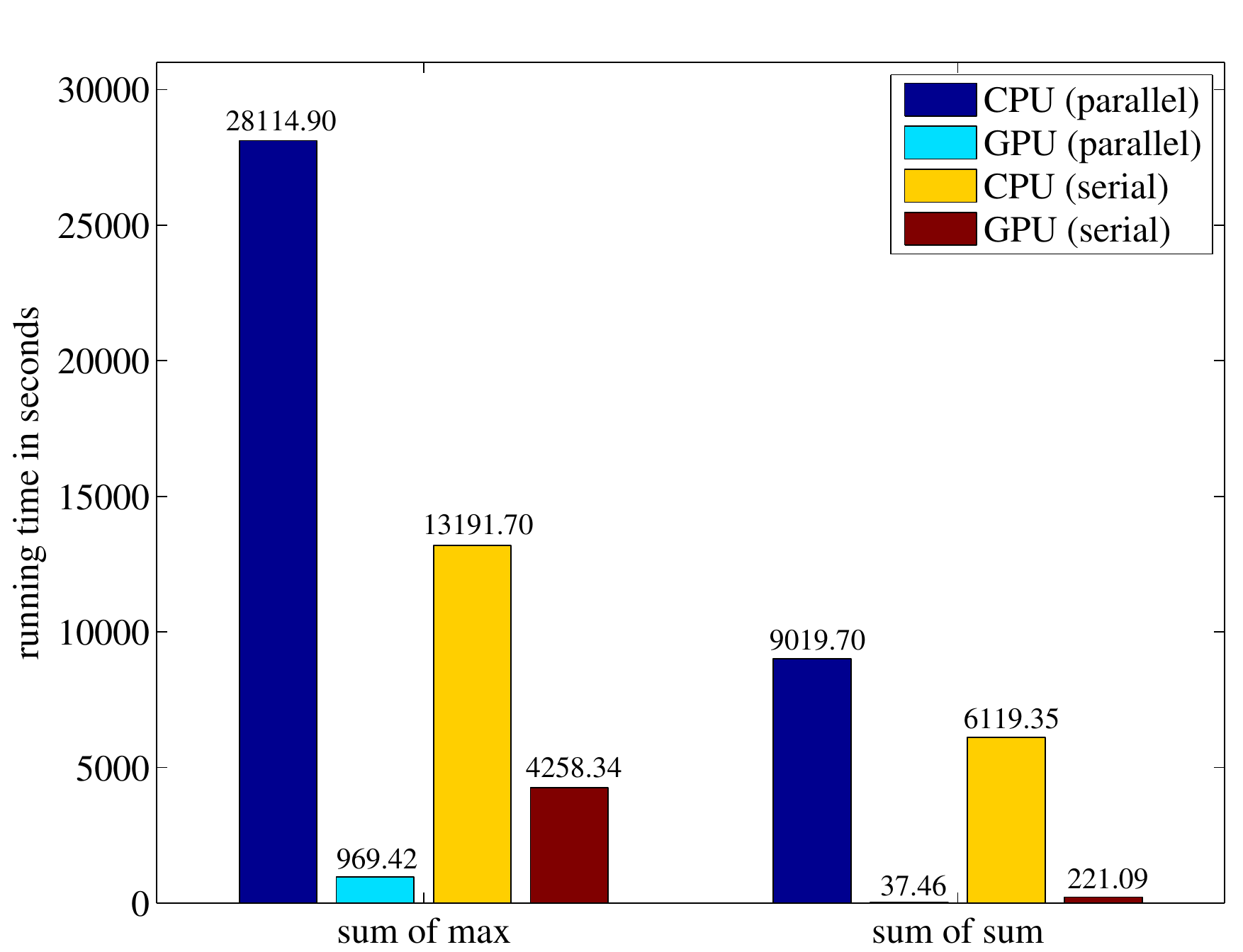}
\caption{Running time in seconds of both rules running on CPU and GPU respectively. We have $C=16$, $L=512$, $50000$ stored messages and $30000$ test messages.}
\label{fig:gpucpu}
\end{figure}

Next, we consider the improvements in runtime achieved by running both rules on a GPU versus on a CPU.
A larger network is simulated in this case.
We have $C=16$ clusters with $L=512$ neurons each, out of which $7$ clusters are erased.
We generate and store $50000$ random messages, and we use a random subset of $30000$ of these to test.
We refer to this experiment setting as Scenario~$2$.
The runtime, in seconds, of both parallel decoding (i.e., decoding a batch of messages concurrently) and serial decoding (i.e., decoding message one after another) on both GPU and CPU are shown in \fig{fig:gpucpu}.

We make three observations:
	First, for each CPU versus GPU and parallel versus serial decoding configuration, \som{} is always significantly slower than \sos{}.
	 For now, let us keep in mind that the fastest retrieval configuration of this entire experiment is roughly $40$ seconds for \sos{} parallel decoding on a GPU.
	We have previously seen that \som{} leads to a much better retrieval accuracy, and so below we focus on achieving the accuracy of the \som{} method while improving its runtime.

	Second, in each group, the bars at the $1$\textsuperscript{st} and $3$\textsuperscript{rd} locations are results for the CPU implementation, and the $2$\textsuperscript{nd} and $4$\textsuperscript{th} bars show results for the GPU implementation.
	Comparing each adjacent pair, we see that the GPU versions consistently run much faster than CPU, as expected.
	The GPU accelerations without any further optimization are respectively (from left-to-right) $29\times$, $3\times$, $240\times$ and $28\times$ faster.

	Finally, parallel decoding is faster than serial decoding on GPU, while the situation reverses on CPU.
	This is reasonable, since parallel decoding can take full advantage of the GPU's computing power.
	However, in the CPU case, if we consider a bundle of $K$ messages, even if only one message does not converge, all $K$ messages will be updated. On the other hand, with serial decoding, the retrieval rule will stop as soon as each individual message converges.

\subsection{Further Accelerating the \som{} Rule}
\begin{figure}
\centering
\includegraphics[scale=.5]{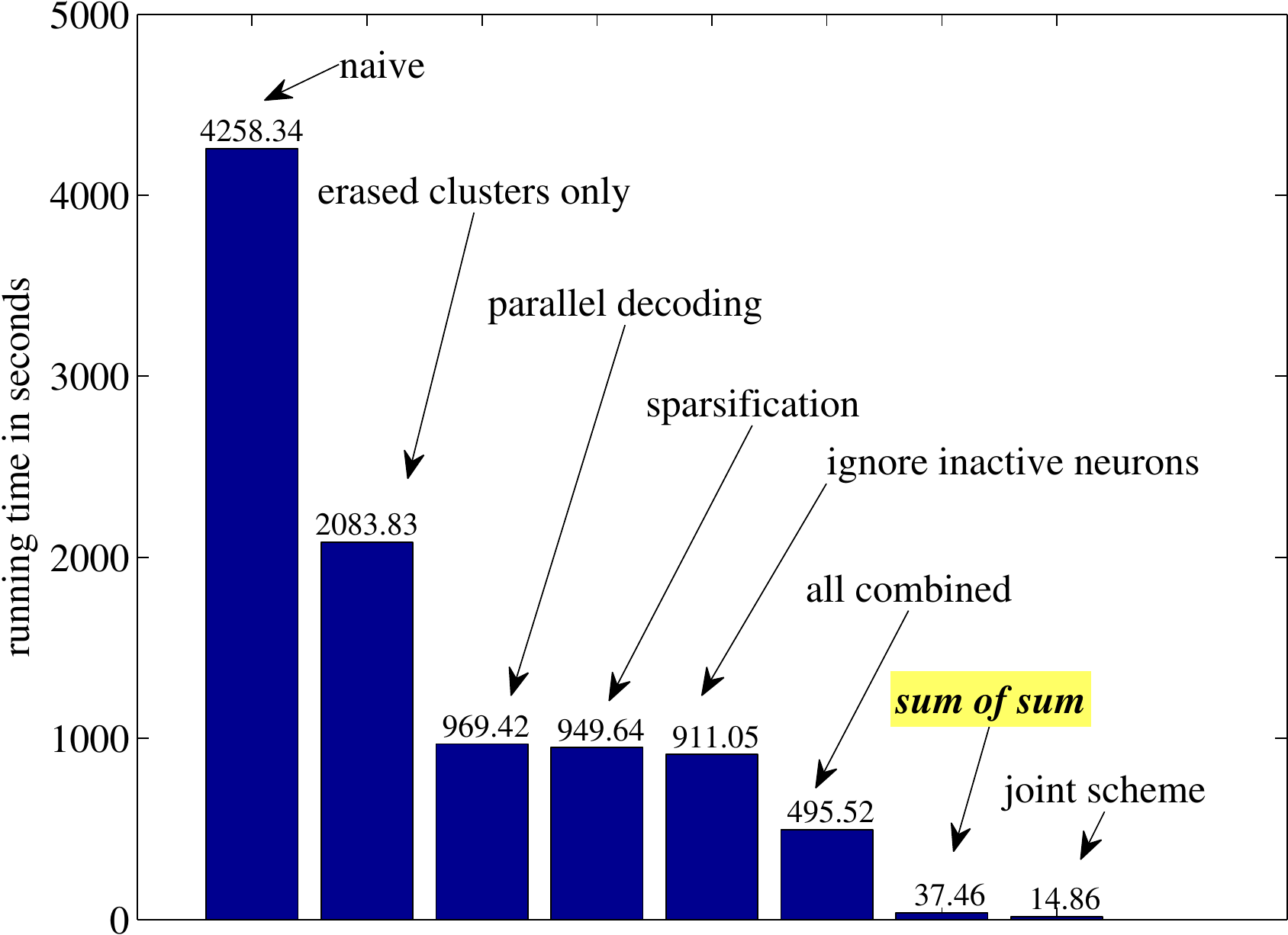}
\caption{Running time in seconds for different acceleration tricks. We have $C=16$, $L=512$, $50000$ stored messages and $30000$ test messages. The first six bars are for \som{}. We plot the performances for different tricks and all combined. We also plot the record of \sos{} and our joint scheme at the end for comparison.}
\label{fig:accelerate}
\end{figure}

In \fig{fig:accelerate} we show the effect of applying the different techniques discussed in Sections~\ref{sec:acceleration} and \ref{sec:property} to accelerate the \som{} rule on a GPU.
Although all of the techniques combined reduce the runtime eightfold, from roughly $4000$ seconds to $500$ seconds, the \som{} rule still cannot compete with \sos{}'s $40$-second spec, which is highlighted in yellow and bold font in the figure. However, the proposed joint scheme cuts the record by another two thirds, achieving the fastest runtime of only $14.86$ seconds for Scenario~$2$.

In \fig{fig:gpucpu}, the faster configuration for \som{} on CPU is the serial decoding scheme, to which we compare, our joint scheme achieves a $880\times$ speedup while retaining the decoding accuracy.

\subsection{Motivation for Combining Rules\label{motivation}}
\begin{figure*}
\centering
\subfloat[running time in seconds for each iteration]{
\includegraphics[scale=0.45]{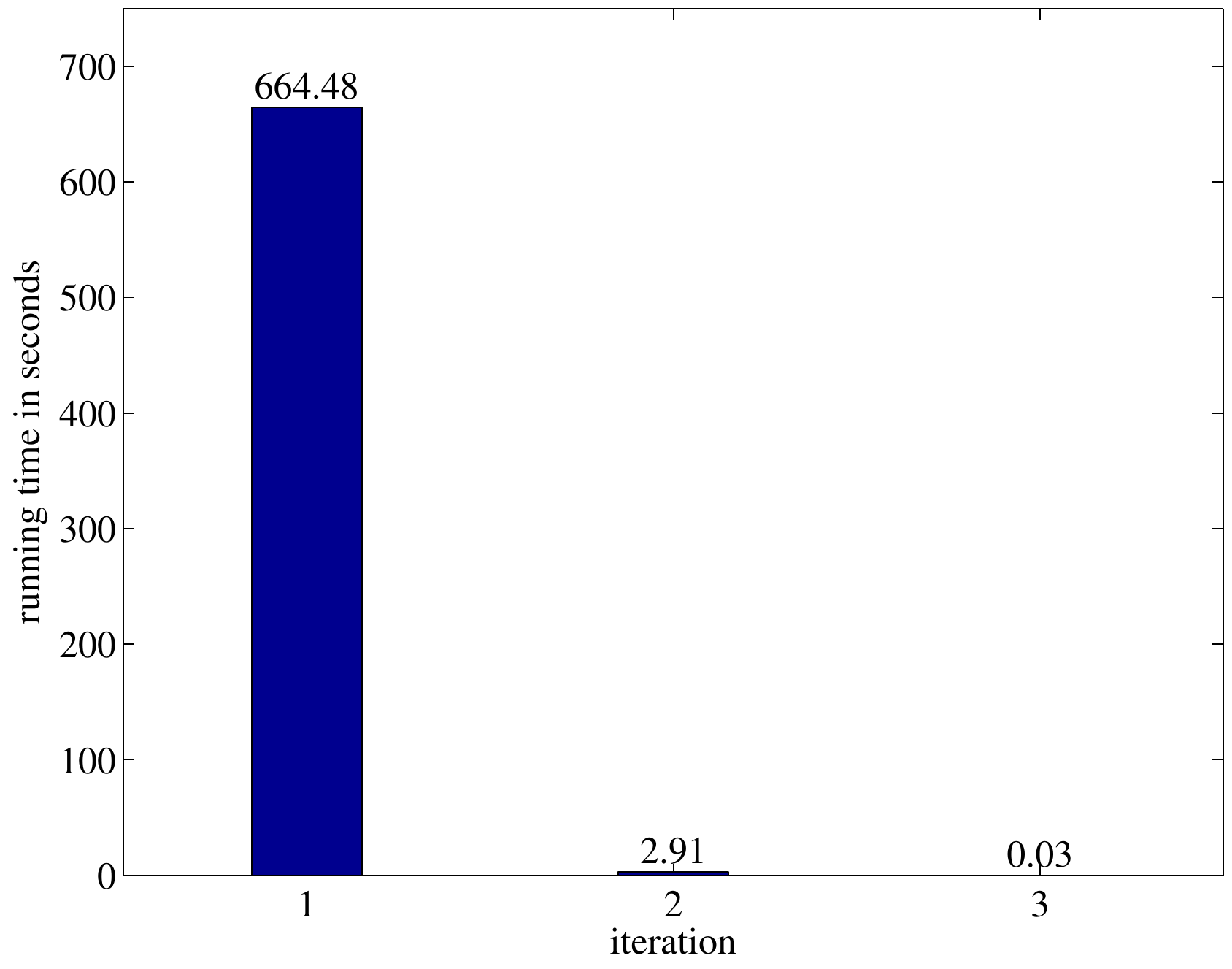}
\label{fig:first}
}
\subfloat[accumulate number of convergent messages after each iteration]{
\includegraphics[scale=0.45]{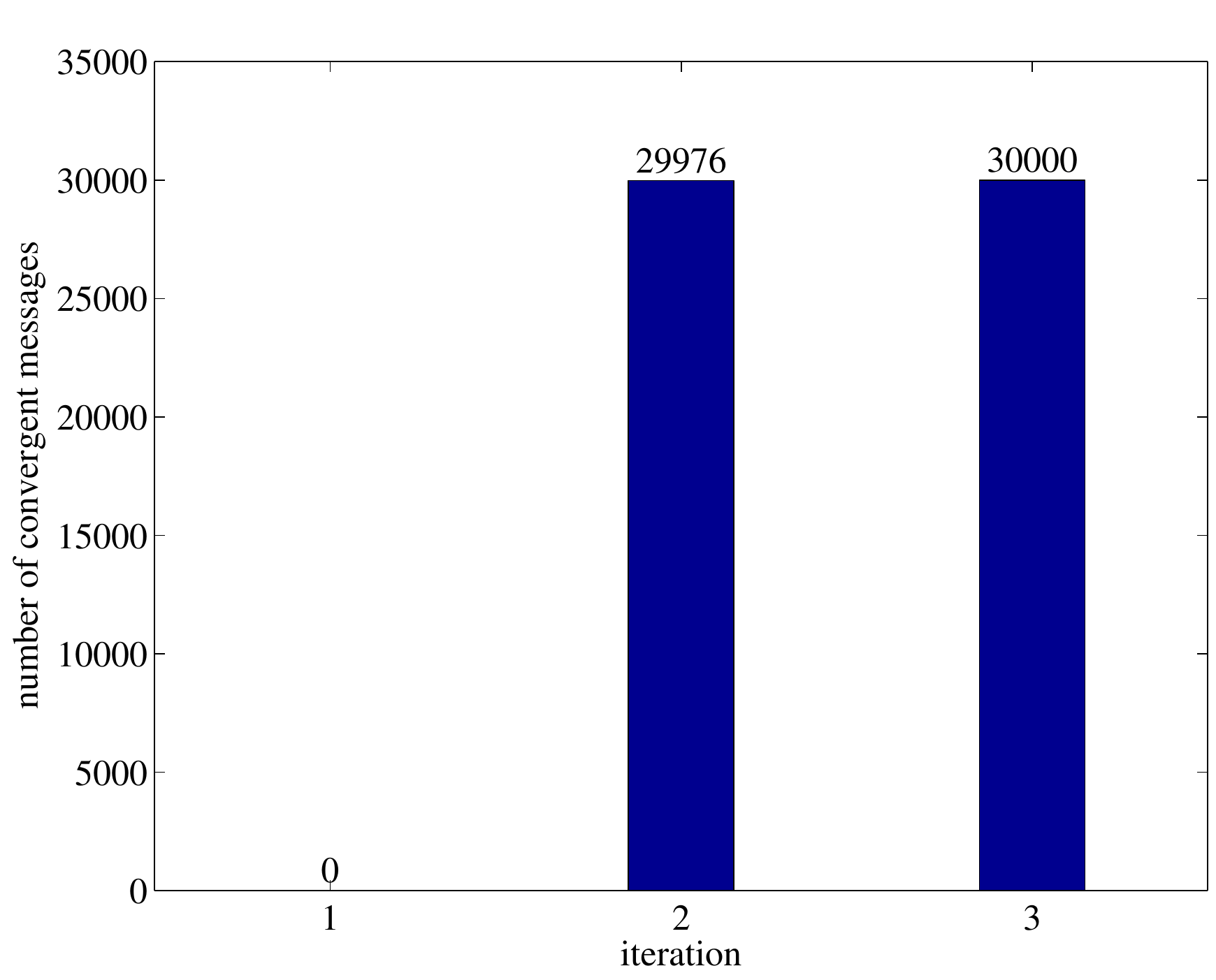}
\label{fig:converge}
}
\caption{A typical run of \som{} with all the neurons in the erased clusters activated initially.
	We experiment Scenario~$2$ where we have $C=16$ clusters, with $L=512$ neurons each, $50000$ messages to memorize and $30000$ messages to test.
	\protect\subref{fig:first} shows the running time in seconds for each iteration.
	\protect\subref{fig:converge} shows the accumulate number of the messages that have converged after each iteration.
}
\label{fig:motivation}
\end{figure*}

Here we provide an example to better illustrate why the joint scheme achieves a significant speedup. We again use Scenario~$2$ and apply all of the acceleration techniques discussed in Section~\ref{sec:acceleration}. We initialize the matrix $V^0$ according to the vanilla \som{}, so that all neurons in clusters corresponding to erased symbols are activated, and only one neuron within each cluster corresponding to a non-erased symbol is active. \fig{fig:motivation} depicts a typical run of the experiment. \fig{fig:first} shows the total runtime spent in each iteration of the \som{} decoding.
One observation is that every subsequent iteration requires less time than its previous one due to the application of Lemma~\ref{thm:dead} and Theorem~\ref{thm:live}; otherwise, the runtime of each iteration should be roughly the same.
Another observation is that the majority of the total runtime is spent in the $1$\textsuperscript{st} iteration; this occurs because initially there are too many unnecessary active neurons in erased clusters, and \som{} demands time to process each one of them.
\fig{fig:converge} shows the number of test messages (out of $30000$) which have converged after each iteration. 

\subsection{New Joint Scheme}
\begin{figure*}
\centering
\subfloat[running time for Scenario~$1$]{
\includegraphics[scale=0.45]{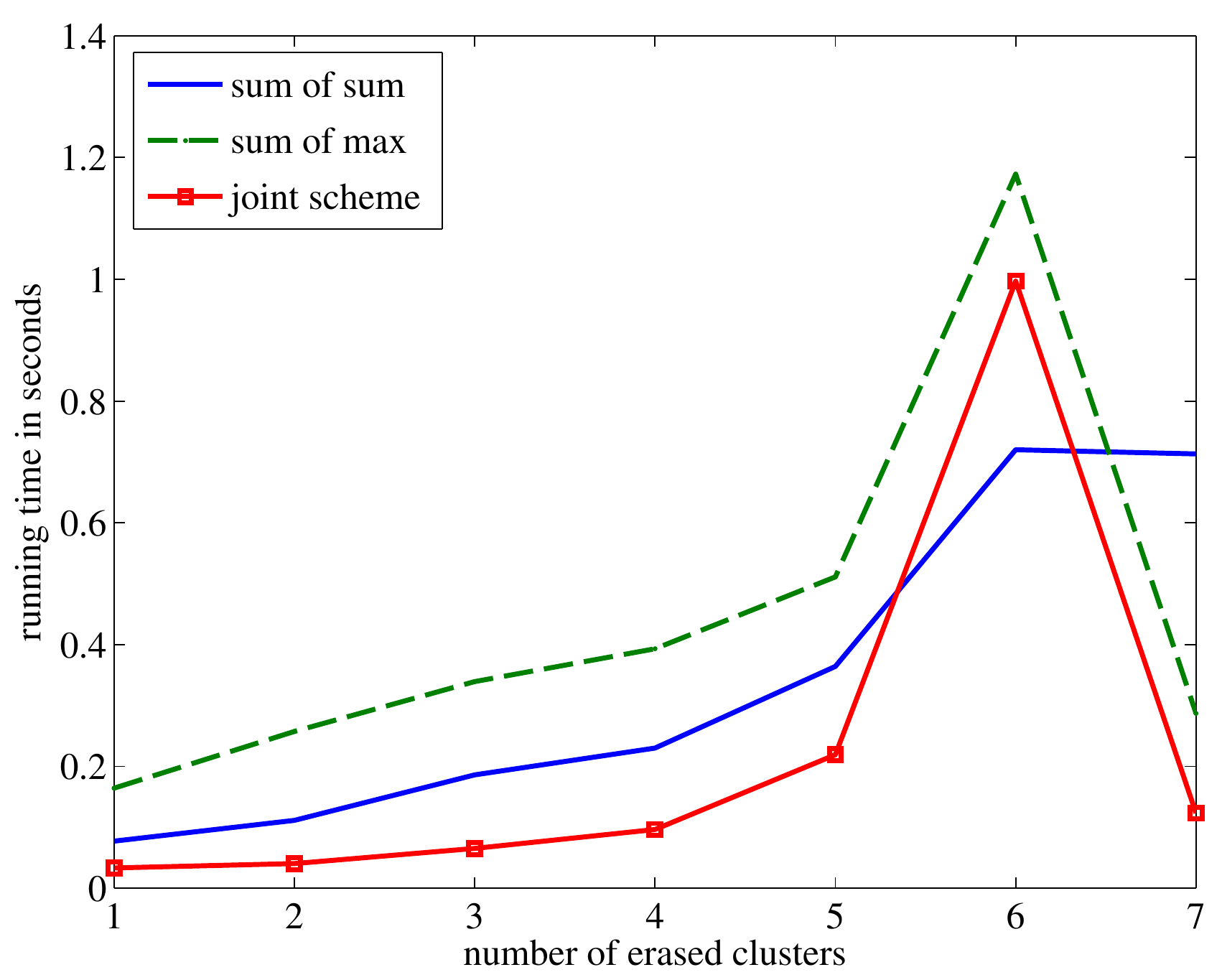}
\label{fig:time1}
}
\subfloat[retrieval rate for Scenario~$1$]{
\includegraphics[scale=0.45]{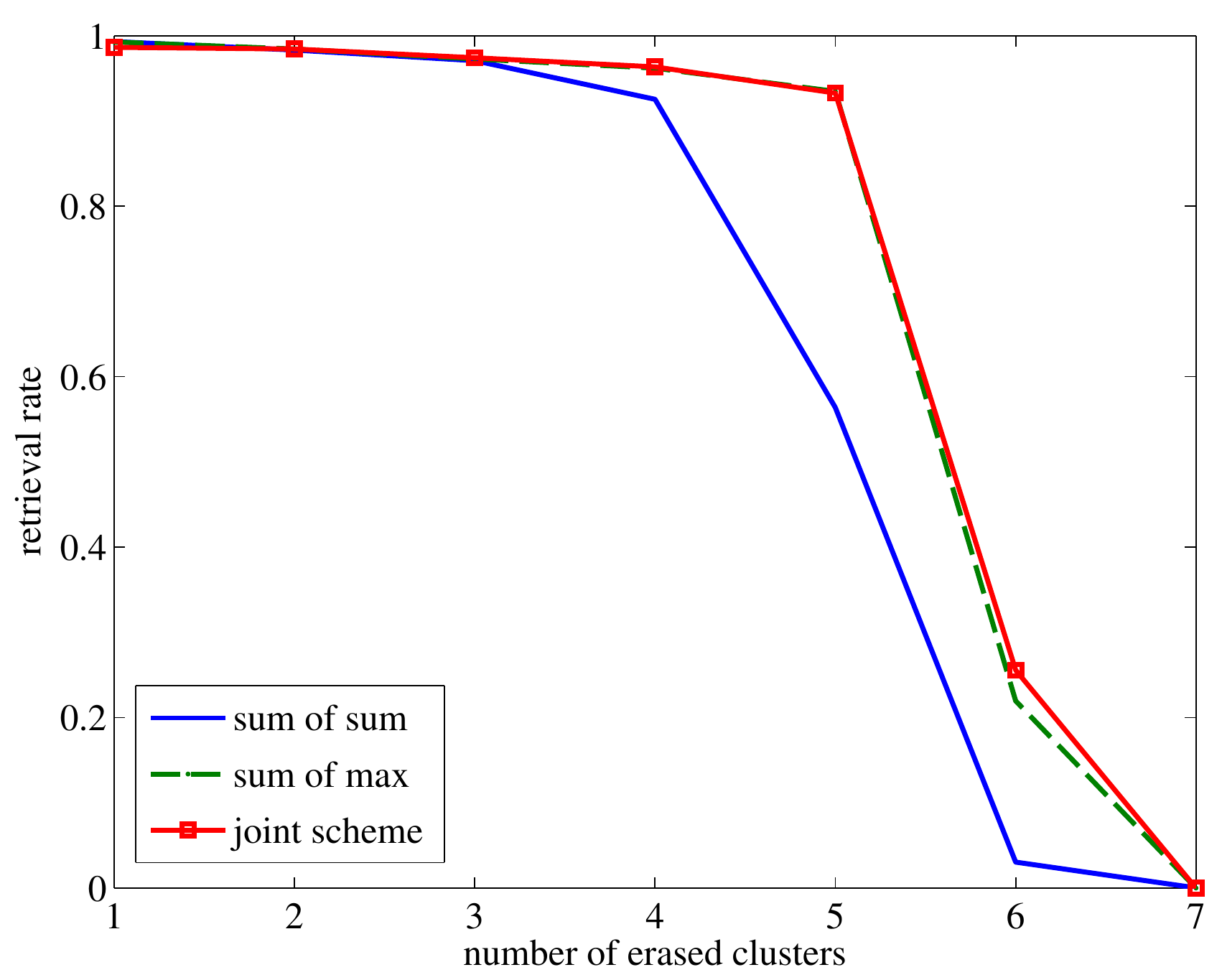}
\label{fig:rate1}
}\\
\subfloat[running time for Scenario~$2$]{
\includegraphics[scale=0.45]{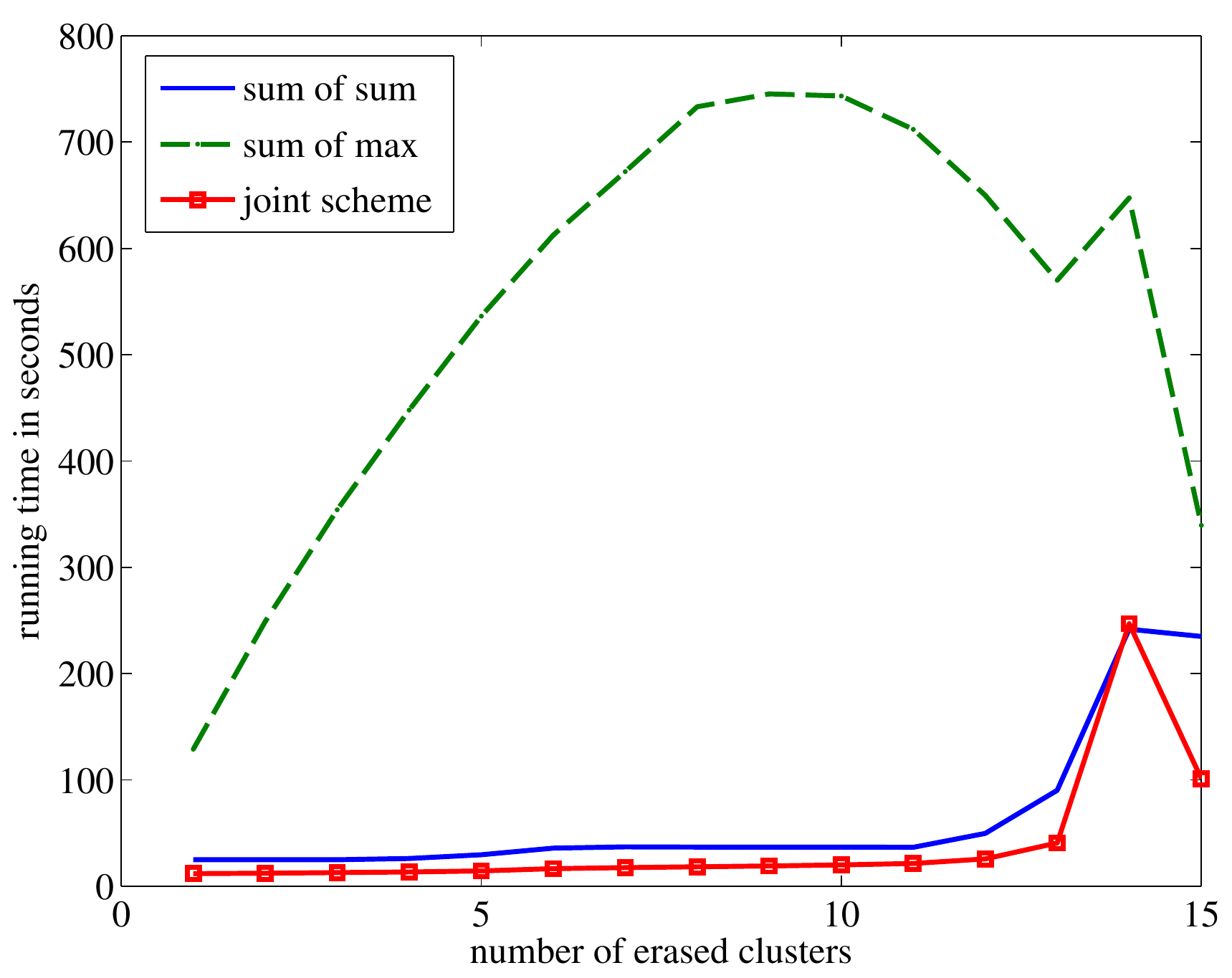}
\label{fig:time2}
}
\subfloat[retrieval rate for Scenario~$2$]{
\includegraphics[scale=0.45]{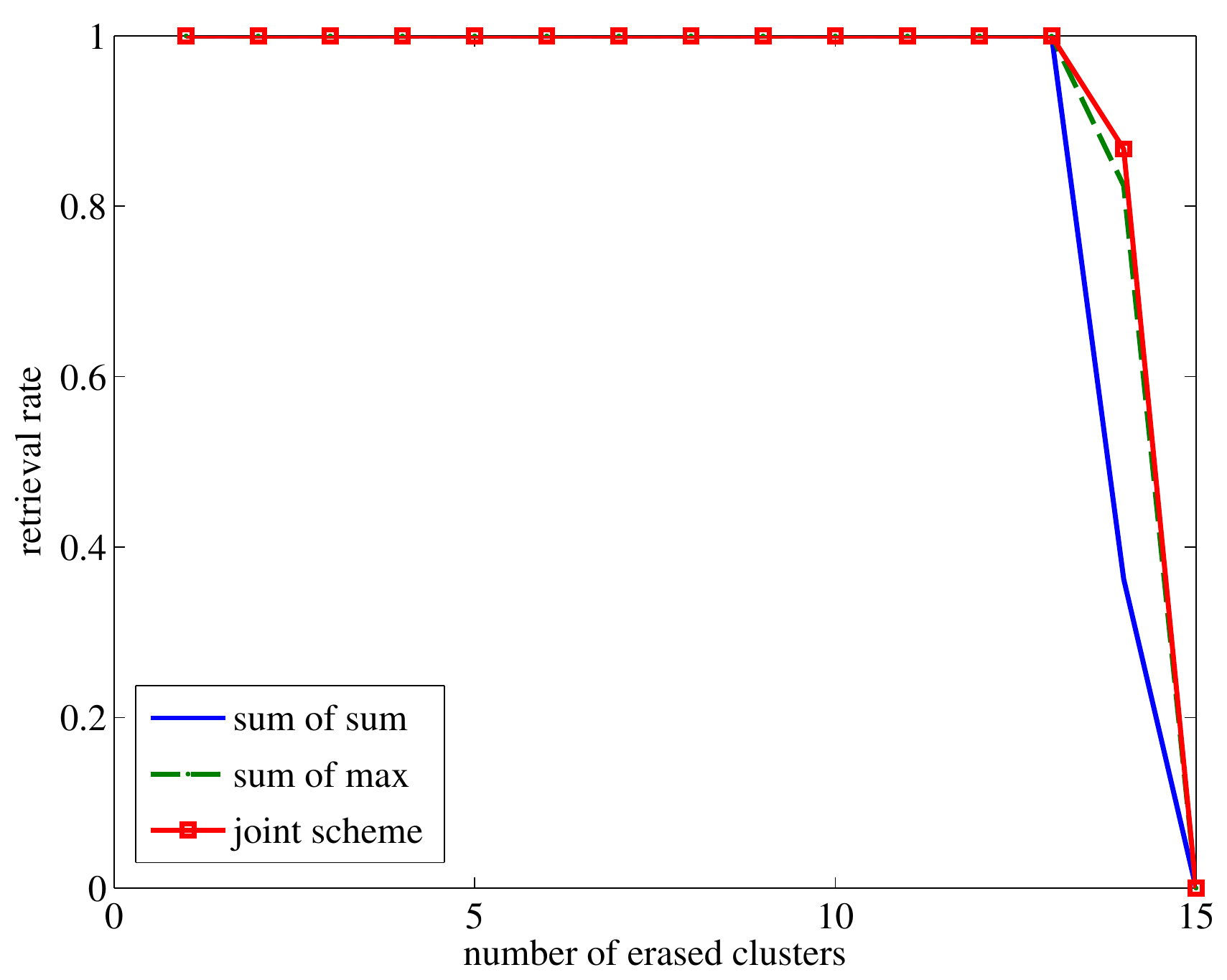}
\label{fig:rate2}
}

\caption{
    The behavior of the joint retrieval scheme in general.
    Both running time in seconds and retrieval rate are plotted respectively as the number of erased clusters increases.
    We set $\gamma=2$ and the maximum number of iterations allowed is $20$.
    \protect\subref{fig:time1} and \protect\subref{fig:rate1} refer to Scenario~$1$ where there are $8$ clusters with $128$ neurons each, $5000$ messages to memorize, and $3000$ messages to test.
    \protect\subref{fig:time2} and \protect\subref{fig:rate2} refer to Scenario~$2$ where there are $16$ clusters with $512$ neurons each, $50000$ messages to memorize, and $30000$ messages to test.
}
\label{fig:joint}
\end{figure*}

Finally, we demonstrate the behavior of the joint decoding scheme across a range of experimental settings.
\fig{fig:joint} shows the runtime (in seconds) and retrieval rate compared with \sos{} and \som{} for both of Scenarios~$1$ and $2$, while varying the number of erased symbols.
The spikes in runtime for \som{} and for the joint scheme in \fig{fig:time1} are due to the fact that decoding becomes more difficult as the number of erased clusters increases, consequently more iterations are required in these cases.
In these settings ($6$ out of $8$ clusters erased for Scenario~$1$, and $14$ out of $16$ clusters erased for Scenario~$2$), although the \sos{} rule is only a bit faster than \som{} and the joint scheme, the retrieval rate is significantly lower.
Another reason that \sos{} runs faster here is due to the limit on the number of iterations which we impose in our experiments. Note that increasing this limit does not improve the retrieval rate, but it can make the runtime arbitrarily worse because \sos{} oscillates.
Also observe that in both \fig{fig:rate1} and \fig{fig:rate2}, the retrieval rates of \som{} and the joint scheme are identical.
In \fig{fig:rate2}, all three approaches achieve effectively a $100\%$ retrieval rate for up to $13$ erased clusters.
This is because the number of messages stored ($50000$) is relatively small for this network.
If this number increases, the deviation in retrieval rate between the joint scheme (as well as \som{}) and \sos{} will be more pronounced.
We conclude from \fig{fig:joint} that the joint retrieval scheme combines the benefits of both existing rules, achieving fast decoding while also maintaining a high retrieval rate.

\subsection{Correlated vs.~Uncorrelated Messages}

The experiments above involve storing and retrieving random messages which are generated uniformly and independently. If messages are correlated and the GBNN architecture is unchanged, we expect the performance to degrade in comparison to the results reported above. Modifying GBNNs to accommodate correlated messages is an area of ongoing investigation. As mentioned in Section~\ref{sec:gbnn}, one approach is to use a ``grandmother layer'', similar to~\cite{knoblauch2005neural}. Another approach is to append each message with i.i.d.~uniform random bits. This requires additional neurons, but reduces the correlation.

\section{Summary\label{sec:summary}}
In this work, we present optimized implementations of the Gripon-Berrou neural network associative memory on a GPU. 
We analyze two existing retrieval rules, namely \sos{} and \som{}.
We show that \sos{} may lead to network oscillation.
However, we manage to prove the convergence of \som{}.
In order to achieve the full speedup, we combine the two rules and propose a hybrid scheme, minimizing the unnecessary computation burdens.
The experimental results show an exciting acceleration against a CPU implementation using an optimized linear algebra library.

GBNNs embrace a LDPC-like sparse encoding setup, which makes the network extremely resilient to noises and errors.
As associative memories serve as building blocks for many machine learning algorithms, we hope the parallel scheme proposed here can be helpful in paving the path to more widespread adoptions of large scale associative memory applications.

An obvious future work is to extend GBNNs to deal with correlated messages, which will severely impair the retrieval performance.
However, the nice structure of GBNNs seems promising in this direction.
In the future, we will try to develop other retrieval schemes, e.g., to handle corrupted patterns as well as incomplete probes.
Since \sos{} runs orders of magnitude faster, another sensible topic is to emulate \som{} using \sos{} so that both performance and speed can be retained simultaneously.
We may also seek the way to generalize GBNN and extend the sparse neural network's use in tasks other than associative memory, e.g., classification and regression.

\bibliographystyle{IEEEtran}
\bibliography{my}

\end{document}